\documentclass[lettersize,journal,twoside]{IEEEtran}

\ifCLASSINFOpdf
  \usepackage[pdftex]{graphicx}
  \graphicspath{{./figures}}
\else
\fi
\usepackage[ruled]{algorithm2e}
\usepackage{caption}
\usepackage{subcaption}
\usepackage{booktabs}
\usepackage{multirow}
\usepackage{dblfloatfix}
\usepackage{amsfonts}
\usepackage{amsmath}
\usepackage{hyperref}
\usepackage{tabularx}

\usepackage{pgfplots}
\usepgfplotslibrary{groupplots}
\pgfplotsset{width=10cm,compat=1.16}
\usepackage{array}
\usepackage{tikz}
\usepackage{xcolor}
\usepackage{orcidlink}

\newcommand{\lightercolor}[3]{
    \colorlet{#3}{#1!#2!white}
}
\newcommand{\darkercolor}[3]{
    \colorlet{#3}{#1!#2!black}
}
\lightercolor{blue}{50}{LightBlue}
\darkercolor{blue}{90}{DarkBlue}
\lightercolor{green}{50}{LightGreen}
\darkercolor{green}{70}{DarkGreen}
\lightercolor{orange}{50}{LightOrange}
\darkercolor{orange}{85}{DarkOrange}
\darkercolor{brown}{85}{DarkBrown}
\lightercolor{brown}{50}{LightBrown}

\begin{document}
%
\title{Persistence of Backdoor-based Watermarks for Neural Networks: A Comprehensive Evaluation}
%
%
%
\author{Anh~Tu~Ngo\orcidlink{0000-0001-7224-6833},
        Chuan~Song~Heng,
        Nandish~Chattopadhyay
        and~Anupam~Chattopadhyay\orcidlink{0000-0002-8818-6983},~\IEEEmembership{Senior~Member,~IEEE,}
\thanks{Anh Tu Ngo, Chuan Song Heng, Nandish Chattopadhyay and Anupam Chattopadhyay are with CCDS, Nanyang Technological University Singapore.}%
}

%
%

\markboth{IEEE Transactions on Neural Networks and Learning Systems}%
{Ngo \MakeLowercase{\textit{et al.}}: Persistence of Backdoor-based Watermarks for Neural Networks: A Comprehensive Evaluation}
%



\maketitle

\begin{abstract}
Deep Neural Networks (DNNs) have gained considerable traction in recent years due to the unparalleled results they gathered. However, the cost behind training such sophisticated models is resource intensive, resulting in many to consider DNNs to be intellectual property (IP) to model owners. In this era of cloud computing, high-performance DNNs are often deployed all over the internet so that people can access them publicly. As such, DNN watermarking schemes, especially backdoor-based watermarks, have been actively developed in recent years to preserve proprietary rights. Nonetheless, there lies much uncertainty on the robustness of existing backdoor watermark schemes, towards both adversarial attacks and unintended means such as fine-tuning neural network models. One reason for this is that no complete guarantee of robustness can be assured in the context of backdoor-based watermark. In this paper, we extensively evaluate the persistence of recent backdoor-based watermarks within neural networks in the scenario of fine-tuning, we propose/develop a novel data-driven idea to restore watermark after fine-tuning without exposing the trigger set. Our empirical results show that by solely introducing training data after fine-tuning, the watermark can be restored if model parameters do not shift dramatically during fine-tuning. Depending on the types of trigger samples used, trigger accuracy can be reinstated to up to 100\%. Our study further explores how the restoration process works using loss landscape visualization, as well as the idea of introducing training data in fine-tuning stage to alleviate watermark vanishing.

\end{abstract}


%
\IEEEpeerreviewmaketitle

\section{Introduction} \label{intro}

\begin{figure}
    \centering
    \includegraphics[scale=0.38]{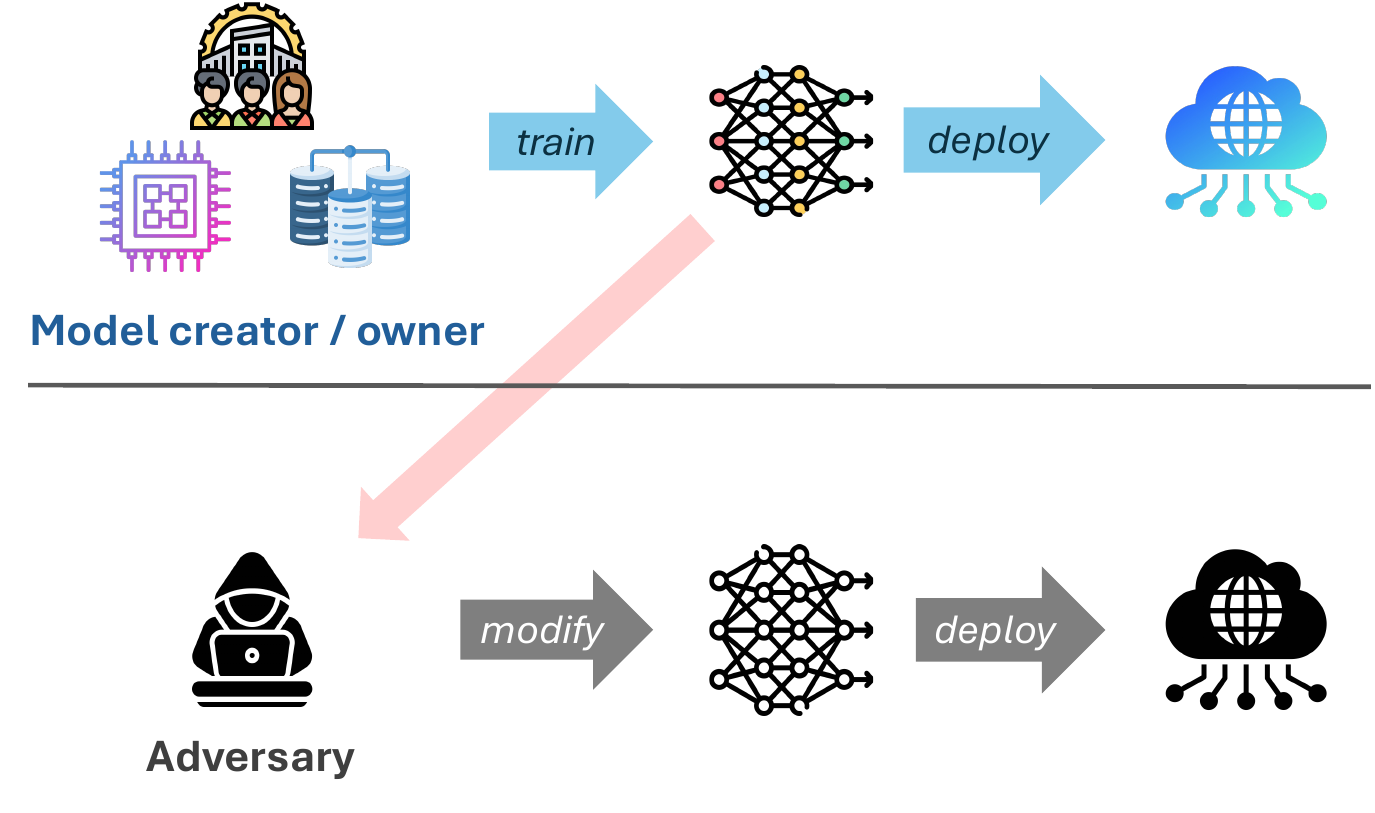}
    \caption{Intellectual property theft of deep neural networks}
    \label{fig:model_steal}
\end{figure}

Recent years have witnessed eminent advancement of Artificial Intelligence (AI) in various aspects of life, ranging from computer vision, natural language processing (NLP) to healthcare. The use of deep neural networks has overshadowed traditional machine learning techniques in those tasks. The phenomenal success of Transformer~\cite{vaswaniAttentionAllYou2017} paved the way to many breakthroughs in language models and even in machine vision. For instance, Transformer-based models such as OpenAI's GPT-3~\cite{brownLanguageModelsAre2020}, GPT-4~\cite{openaiGPT4TechnicalReport2024}, Google's LaMDA~\cite{thoppilanLaMDALanguageModels2022} and PaLM 2~\cite{anilPaLM2Technical2023} have become the dominant large language models (LLMs) and now serve as backbones in ChatGPT and Bard chatbots. Nonetheless, these models usually consist of up to hundreds of billions of parameters and it costs millions of dollars to train them. Aside from training cost, the infrastructure, data acquisition and human resource payment can make up colossal expense for the host companies. Such exorbitant cost and enormous effort have made these models valuable intellectual properties (IP) of the companies. Furthermore, with the growth of machine learning as a service (MLaaS)~\cite{ribeiroMLaaSMachineLearning2015}, the privacy of machine learning models is exposed to various threats. For instance, the information of a model can be leaked to adversaries via malware infection or insider attacks. The unauthorized parties then make modifications to the cloned model so that it differs from the original one and deploy it on their own service, as shown in Figure~\ref{fig:model_steal}. To that end, sufficient care for model's privacy should be taken into consideration.

One of the effective techniques to guard DNNs from illegal usage is watermarking, in which some special patterns are embedded into the host documents. Watermark has been used for a long time in digital documents like photos, videos, sounds, etc,. In the past decade, researchers have adopted the idea of watermarking to machine learning models, especially DNNs. The first work in DNN watermark is from~\cite{uchidaEmbeddingWatermarksDeep2017}, whose idea is inspired by conventional digital watermark techniques that embed hidden signature into the model's parameters by modifying the regularizer. More recent DNN watermarking techniques are based on the idea of backdoor, which is first proposed by Adi et al.~\cite{adiTurningYourWeakness2018}. The idea is to train the DNNs so that they output specific predictions for a specifically designed dataset.

In real-world use cases, it is common that model owners create their own pretrained DNNs and distribute them publicly, or to subscribed clients. In those situations, DNNs do not invariably stay static. Instead, model users usually make changes to the DNNs so that they suit better to their particular domain by transfer learning or fine-tuning, in which model's parameters are modified by being trained on newer data. This process challenges the persistence and robustness of backdoor watermarks as the they are embedded in model's parameters. Recent work from~\cite{shafieinejadRobustnessBackdoorbasedWatermarking2021,chenREFITUnifiedWatermark2021a} aims at removing backdoor watermarks via fine-tuning. They demonstrate that most DNN backdoor watermarks are vulnerable to removal attack like fine-tuning, although~\cite{adiTurningYourWeakness2018,liuFinePruningDefendingBackdooring2018a} claim that fine-tuning is not sufficient to remove backdoor watermarks. Indeed, it is crucial to study the effects of fine-tuning on watermark persistence because fine-tuning is one of the most straightforward techniques to manipulate model parameters. From an adversary's perspective, fine-tuning is the most feasible approach they can perform to remove a watermark from a DNN. From the standpoint of a model user, fine-tuning is beneficial for their business as it tailors the models to meet their needs. In both cases, the DNN watermark is at risk of erosion.

To this end, we focus on evaluating the persistence of existed backdoor watermark schemes against fine-tuning, as well as how to enhance their resilience in such events. Our contributions can be summarized as follows:
\begin{itemize}
    \item Watermark restoration: we propose a data-driven method utilizing the idea of basins of attraction of local minima. Our experiments show this method helps regain the watermark accuracy after fine-tuning process, which involves retraining the DNN with the original clean training set without further exposure of trigger samples to the model. To the best of our knowledge, this is the first work that introduces the idea of retraining for watermark restoration. Although most of the current research focuses on improving watermark embedding schemes, we believe our new approach is important to boost the watermark performance when it gets weakened.
    \item Loss landscape analysis: we analyze the optimization trajectory of model parameters using loss landscape geometry. The analysis illustrates how model parameters traverse the landscape during retraining and how the landscape looks like with respect to particular trigger set types. We found these visualizations to be very useful to understand how the phenomenon of watermark restoration happens.
    \item Fine-tuning with mitigated watermark degradation: based on the idea of restoring watermark with retraining, we experiment with blending original training data into fine-tuning stage to investigate its effectiveness in reducing watermark vanishing. This concept is useful, especially in the scenario that authorized model users fine-tune a watermarked model with their own data. If they fine-tune with a mix of original training data, it is quite likely that watermark erosion will be alleviated.
\end{itemize}

Regarding watermark restoration, this intriguing property of local minima allows model owners to bring back (part of) watermark behavior of a suspiciously public model, leading to successful ownership verification to claim the IP right of that model. Meanwhile, the idea of fine-tuning with blending data can be useful in cases when authorized model users want to perform fine-tuning on their own, without the need to send their proprietary dataset to model owner via API access. On the one hand, this fine-tuning scheme ensures that any undesirable leakage of users' data is preventable. On the other hand, model owners, with the proposed fine-tuning technique, are still able to alleviate the watermark vanishing without risking the secrecy of trigger samples during that process. We detail these two concepts later in Section~\ref{section:theory}.

The paper is organized as follows, Section~\ref{background} reviews the background of watermark requirements and related work for DNN watermarking, Section~\ref{section:theory} details the threat model and and the theories behind our proposed methodology, Section~\ref{experiment} shows our experiments with results, and Section~\ref{section:conclusion} concludes the paper.

\section{Background \& Related Work} \label{background}
DNN watermarks, though differ in terms of mechanism compared to digital watermarks, need to fulfill some requirements to successfully protect the model's privacy. In the following, we discuss the fundamental requirements for DNN watermarks. Then we review the related work on neural network backdoors and some notable research attempts to turn malicious backdoors into privacy guards in neural networks.

\subsection{Requirements for DNN watermarks}
\begin{figure}
    \centering
    \begin{tikzpicture}[>=stealth,line width=2.5pt,n/.style={text=white,fill=#1,midway},nodes={font=\bfseries\sffamily}]
    \renewcommand{\baselinestretch}{.8} 
    \def\r{3} 
    \colorlet{colorA}{red}
    \colorlet{colorB}{teal}
    \colorlet{colorC}{green!50!black}
    
    \path[nodes={align=center}]
    (0,0) node{Watermark\\Requirement\\Trilemma}
    (90:\r) node (C) {
    \includegraphics[scale=0.05]{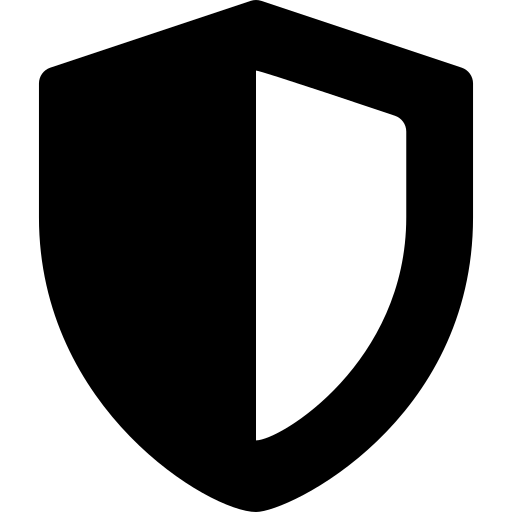}
    }
    (210:\r) node (B) {
    \includegraphics[scale=1.5]{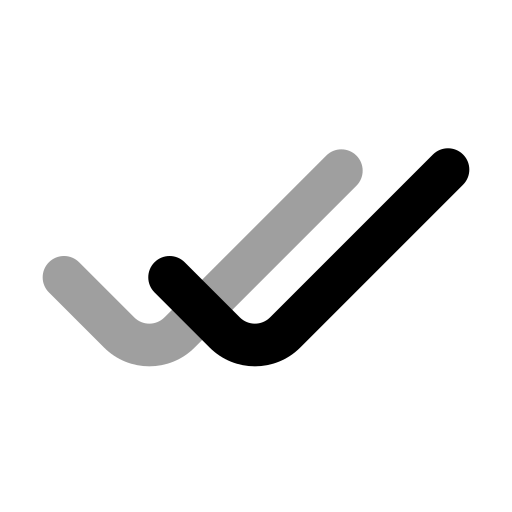}
    }
    (-30:\r) node (A) {
    \includegraphics[scale=0.07]{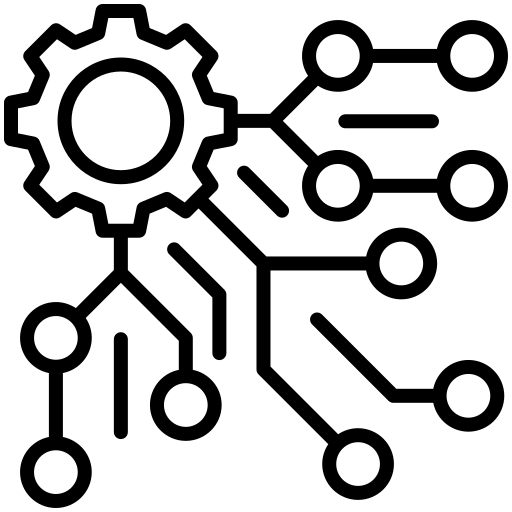}
    }
    ;
    \draw[<->,colorC] (A)--(B);
    \draw[<->,colorA] (B)--(C);
    \draw[<->,colorB] (C)--(A);
    \node[below left] at (-2.2,-1.9) {Fidelity};
    \node[below right] at (2.2,-1.9) {Capacity};
    \node[above] at (0,3.3) {Persistence};
    \end{tikzpicture}   
    \caption{Trilemma between persistence, capacity and fidelity}
    \label{fig:triangle}
\end{figure}
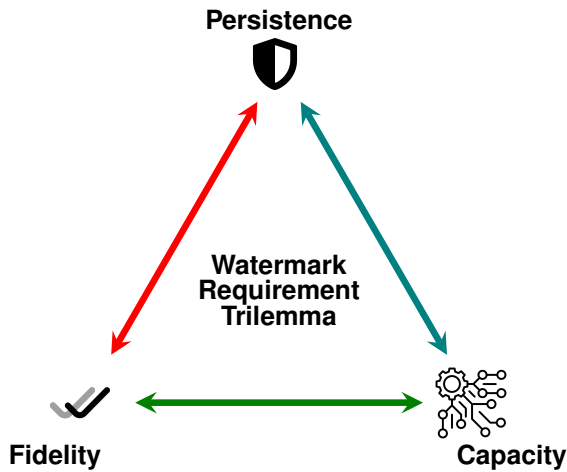

The primary difference between conventional digital watermarking and DNN watermarking is that for digital documents, the watermark can be injected directly into the host documents. Whereas in the case of DNN, the watermark cannot be directly embedded into the model weights. The watermark embedding process must happen during training. Despite this, both digital watermarking and DNN watermarking have to satisfy a good trade-off between persistence, capacity and fidelity (for DNN watermark) or imperceptibility (for digital watermarking). This trade-off can be viewed as a triangle in Figure~\ref{fig:triangle}, which is discussed in~\cite{liSurveyDeepNeural2021}. However, one downside of this watermarking scheme is that it requires explicit inspection of model parameters in order to verify the ownership. In what follows, we briefly review the key requirements for DNN watermarks, which are also mentioned in~\cite{uchidaEmbeddingWatermarksDeep2017,darvishrouhaniDeepSignsEndtoEndWatermarking2019}.

\textbf{Persistence.} This is the ability of a watermark to be retained from the host documents. In the context of DNN watermarking, the presence of watermark footprint should be preserved to a great extent in the event of model manipulations, e.g. fine-tuning, model compression. In other words, this property is the robustness of watermark against model attacks/modifications.

\textbf{Fidelity.} As regards digital watermarking, fidelity is considered equivalent to \emph{imperceptibility}, in which watermarks should not degrade the quality of host documents. In terms of DNN watermarking, a good fidelity means that the watermark does not have a great detrimental impact on the model performance on its original task.

\textbf{Capacity.} This is the amount of information that can be embedded into host contents, expressed as the number of bits or payload. Most common DNN watermark schemes are either zero-bit or multi-bit watermarks.

There are a few more requirements that a DNN watermarking scheme should satisfy to be considered of good quality. Table~\ref{table:wmreq} summarizes the most common criteria for assessing the quality of a DNN watermark scheme.

\begin{table}
    \centering
    \begin{tabular}{l p{6.3cm}}
        \toprule
        \textbf{Criterion} & \textbf{Description}\\
        \midrule
        Persistence & The watermark should resist various attacks and model modifications \\
        \midrule
        Capacity & The capability of embedding large amount of information into host neural network\\
        \midrule
        Fidelity & The watermark should not significantly affect the performance of target NN on original task\\
        \midrule
        Integrity & The false alarm rate (or number of false positives) should be minimal\\
        \midrule
        Security & The presence of watermark should be secret and undetectable\\
        \midrule
        Efficiency & The computational overhead of watermark construction and verification should be negligible\\
        \midrule
        Generality & The watermark technique can be adaptive to various models, datasets and learning tasks\\
        \bottomrule
    \end{tabular}
    \caption{Requirements for DNN watermarks}
    \label{table:wmreq}
\end{table}

\subsection{Related Work}
The first work on protecting the IP of neural networks using watermark was proposed by Uchida et al.~\cite{uchidaEmbeddingWatermarksDeep2017}. In this work, the secret key is a specially designed vector $X$ with $T$-bit length. The watermark embedding occurs during model training and is done by adding an \emph{embedding regularizer} term to the original loss function. This can be written as $E(w)=E_0(w) + \lambda E_R(w)$ where $E_0(w)$ is the loss for original task and $E_R(w)$ is the additional embedding regularizer imposing a statistical bias on the parameters $w$. To extract and verify the watermark, model owners simply have to compute the project of $w$ onto $X$, which indicates the presence of watermark by comparing with a pre-defined threshold.

\textbf{Backdoor in neural networks.} After the work in~\cite{uchidaEmbeddingWatermarksDeep2017}, many researchers have been actively tackling DNN watermaking problem in various ways. A prominent technique to embed a watermark into neural networks is backdooring. According to~\cite{guBadNetsEvaluatingBackdooring2019}, backdooring in neural networks corresponds to the process of training a neural network in such a way that it outputs wrong labels for certain input samples and is regarded as one kind of \emph{data poisoning}. The power of modern DNN models stems from their over-parameterization, that is, the number of model parameters is much more than the number of training samples so that the models have more capability to solve their original task. However, this characteristic paves the way for backdooring, hence a security weakness in DNN models. Attacks of this kind are often referred to as \textit{poisoning-based}, the most common class of backdoor attacks. Work from Li et al.~\cite{liBackdoorLearningSurvey2024} is among the most comprehensive surveys about DNN backdooring, in which a wide variety of techniques and scenarios are discussed.

\textbf{Backdoor-based watermarking.} Traditionally, backdoors are considered undesirable to AI security. Nevertheless, Adi et al.~\cite{adiTurningYourWeakness2018} turned this ``badness'' into a ``privacy guard'' by using backdoor as a secret key to claim ownership. The authors designed a trigger dataset comprising abstract images which are completely unrelated to the primary task and are randomly mislabeled. The watermark embedding can be done either during fine-tuning or training from scratch via data poisoning. Zhang et al.~\cite{zhangProtectingIntellectualProperty2018} proposed various approaches to generate trigger samples, i.e. embedding a text onto the image, perturbing the image with Gaussian noise or using out-of-distribution samples. Rouhani et al.~\cite{darvishrouhaniDeepSignsEndtoEndWatermarking2019} proposed DeepSigns, which introduces a hybrid method to embed a watermark into a DNN. There are two key steps in this framework: First, a $N$-bit string $b$ is embedded into intermediate layers of the target NN and the loss function is modified with additional regularizers that enforce the hidden layers' activation to fit better to a Gaussian distribution and embed the watermark string $b$ via projection, similar to that of~\cite{uchidaEmbeddingWatermarksDeep2017}. Second, DeepSigns watermarks the network's output layer by selecting watermark keys, or input samples, whose activation lies in the \emph{rarely explored} area of intermediate layers. In other words, the neural network produces incorrect predictions for these samples. The target network is then fine-tuned to classify these samples correctly.

A few studies make use of adversarial examples to generate trigger data. Frontier Stitching~\cite{lemerrerAdversarialFrontierStitching2020} is the first watermark scheme to leverage adversarial examples. In this scheme, the trigger data contains \emph{true adversaries}, which are perturbed samples that fool the model into outputting incorrect predictions. It also includes \emph{false adversaries}, where adversarial noises are added to the original samples so that the classification results are not affected. Both types of adversarial samples are generated in such a way that they are close to the decision boundaries (frontiers), which ensures that the decision frontiers of watermarked classifier do not deviate drastically from the non-watermarked one's. ROWBACK~\cite{chattopadhyayROWBACKRObustWatermarking2021} also adopts adversarial examples for trigger set generation but differs in the labeling procedure. The algorithm employs FGSM~\cite{goodfellowExplainingHarnessingAdversarial2015} to create adversarial samples. Another novel contribution of ROWBACK is uniform watermark distribution, which means the watermark footprint exists across the whole NN's layers. To achieve that, the choice of which layers to be unfrozen during training changes every iteration.

Most of the backdoor watermark schemes claim their robustness to removal attacks with little theoretical guarantee. Recently, Bansal et al.~\cite{bansalCertifiedNeuralNetwork2022} proposed a certifiable trigger-based watermark that utilizes randomized smoothing to enhance robustness. The paper theoretically shows that certification guarantees watermark's robustness within a $l_2$-norm ball of parameters modification. Ren et al.~\cite{renDimensionindependentCertifiedNeural2023a} proposed a novel smoothing technique based on mollifier theory which achieves a certified watermark robustness against $l_p$-removal attacks with large $p$. Jia et al.~\cite{jiaEntangledWatermarksDefense2021} addressed a fundamental limitation of previous watermark strategies, i.e. watermark task is learned separately from the primary classification task. This work proposed entangled watermark embedding (EWE), which adds some entanglement between the representations of both tasks. With this scheme, an adversary is forced to significantly sacrifice the performance on primary task if they want to remove the watermark. Gan et al.~\cite{ganRobustModelWatermark2023} discovered that there exist many watermark-removed models in the vicinity of original marked model in parameter space and introduced a minimax method to recover the watermark behavior of these models. Wang et al.~\cite{wangFreeFinetuningPlugPlay2023} proposed a novel watermark mechanism that injects a proprietary model into the target model. This approach, according to the authors, allows the target model to achieve desirable main task performance without sacrificing its capacity for watermark classification task due to unchanged target model parameters. CosWM from Charette et al.~\cite{charetteCosineModelWatermarking2022} is a watermark scheme resistant to ensemble distillation. The core technique behind it involves adding some periodic perturbation to model's output. The authors showed that the cosine perturbation is difficult to remove via outputs averaging during ensemble distillation. Li et al.~\cite{liUntargetedBackdoorWatermark2022} introduced an untargeted backdoor watermark (UBW) scheme, which differs from other backdoor-based methods in the model's behavior against backdoored samples. Compared to previous backdoor watermarks, model's predictions on backdoor samples are non-deterministic, which makes it harder for adversaries to manipulate the model's behavior. \cite{huaUnambiguousHighFidelityBackdoor2024} analyzes the vulnerability of most backdoor watermarks to ambiguity attack and proposed an unambiguous backdoor-based watermark scheme which is robust to this attack. Apart from watermarking classification models, some works extend the application of watermark to image processing~\cite{quanWatermarkingDeepNeural2021} or object detection~\cite{lansariBlackBoxWatermarkingModulation2024}.

\section{Local Minima and Watermark Restoration}\label{section:theory}
\subsection{Preliminaries}\label{section:preliminaries}

\textbf{Watermark embedding.} In this work, we focus on fine-tuning watermarked neural networks. A model owner trains their model $f_{\theta}$, where $\theta \in \mathbb{R}^N$ is model parameters, on clean dataset $\mathcal{D}_\textsc{Train}$ to perform a specific classification task $\mathcal{T}$. To verify ownership of the model, a set of trigger samples $\mathcal{D}_{\text{WM}}$ are embedded into $f_\theta$ during training and extracted during verification process. To achieve this goal, the optimization process tries to solve:

\begin{equation}\label{eq:optimization}
    \theta^{*} = \arg\min_{\theta}\mathcal{L}(y, f_\theta(x))
\end{equation}
where $x\in \mathcal{D}_\textsc{Train} \cup \mathcal{D}_{\text{WM}}$. Here, we assume the watermark is embedded during pretraining phase and samples from $\mathcal{D}_\textsc{Train}$ and $\mathcal{D}_{\text{WM}}$ significantly differ in probability distribution, which is either $x_\textsc{Train} \overset{d}{\neq} x_{\text{WM}}$ or $y_\textsc{Train} \overset{d}{\neq} y_{\text{WM}}$. This optimization ensures that a basic watermark scheme must at least satisfy two crucial properties: 
\begin{enumerate}
    \item \textit{Functional preserving:} the ability of watermarked model to achieve comparable performance to non-watermarked model on the main classification task\\ $Pr(y_\textsc{Train}=f_\theta(x_\textsc{Train})) \approx Pr(y_\textsc{Train}=f'(x_\textsc{Train}))$.
    \item \textit{Verifiability:} the watermarked model must be clearly distinguishable from its non-watermarked variant by their performance on trigger data, which helps model owner claim the ownership. Typically, a non-marked model performs very poorly on this dataset while marked model gives a very good performance. \\$Pr(y_{\text{WM}}=f_\theta(x_{\text{WM}})) \gg Pr(y_{\text{WM}}=f'(x_{\text{WM}}))$.
\end{enumerate} where $f_\theta$ and $f'$ denote watermarked model and non-watermarked model respectively.

\textbf{Model fine-tuning.} In real-world scenarios, it is quite common that training data $\mathcal{D}_\textsc{Train}$ and fine-tuning data $\mathcal{D}_\textsc{Finetune}$ come from different distributions of different domains, which requires reconfiguring a few last layers of the NN. However, in the scope of this paper, we consider the scenario that $\mathcal{D}_\textsc{Train}$ and $\mathcal{D}_\textsc{Finetune}$ are similar in terms of domain, so that the watermark evaluation can be done throughout all layers of the NNs.

\begin{figure}
    \centering
    \includegraphics[width=.48\textwidth]{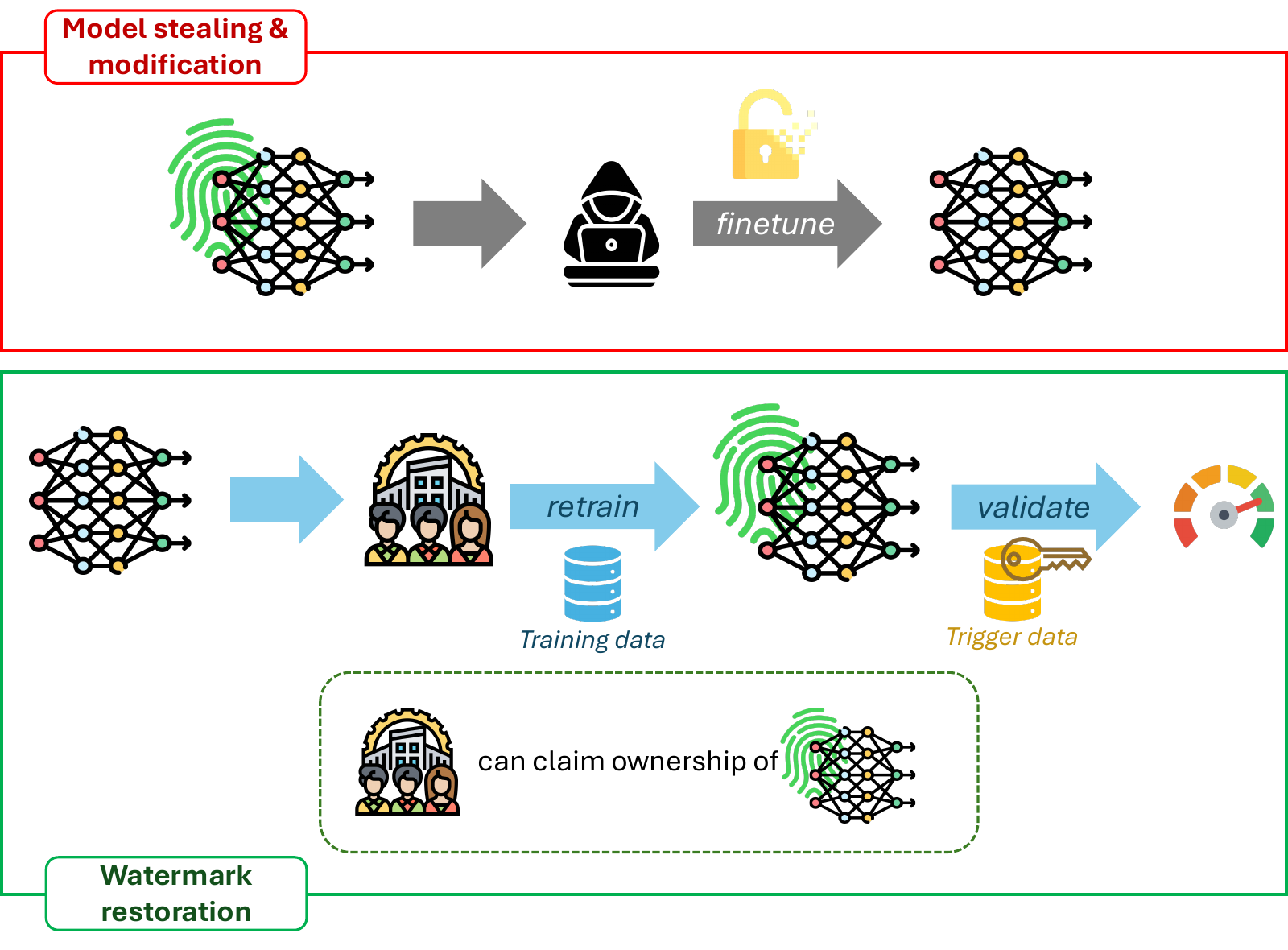}
    \caption{Watermark restoration after fine-tuning attack - the model owner retrains the model with original training data and evaluates the accuracy on trigger samples. The intuition is only the marked model has improved trigger accuracy after retraining process.}
    \label{fig:threatmodel_retrain}
\end{figure}

\begin{figure}
    \centering
    \includegraphics[width=.48\textwidth]{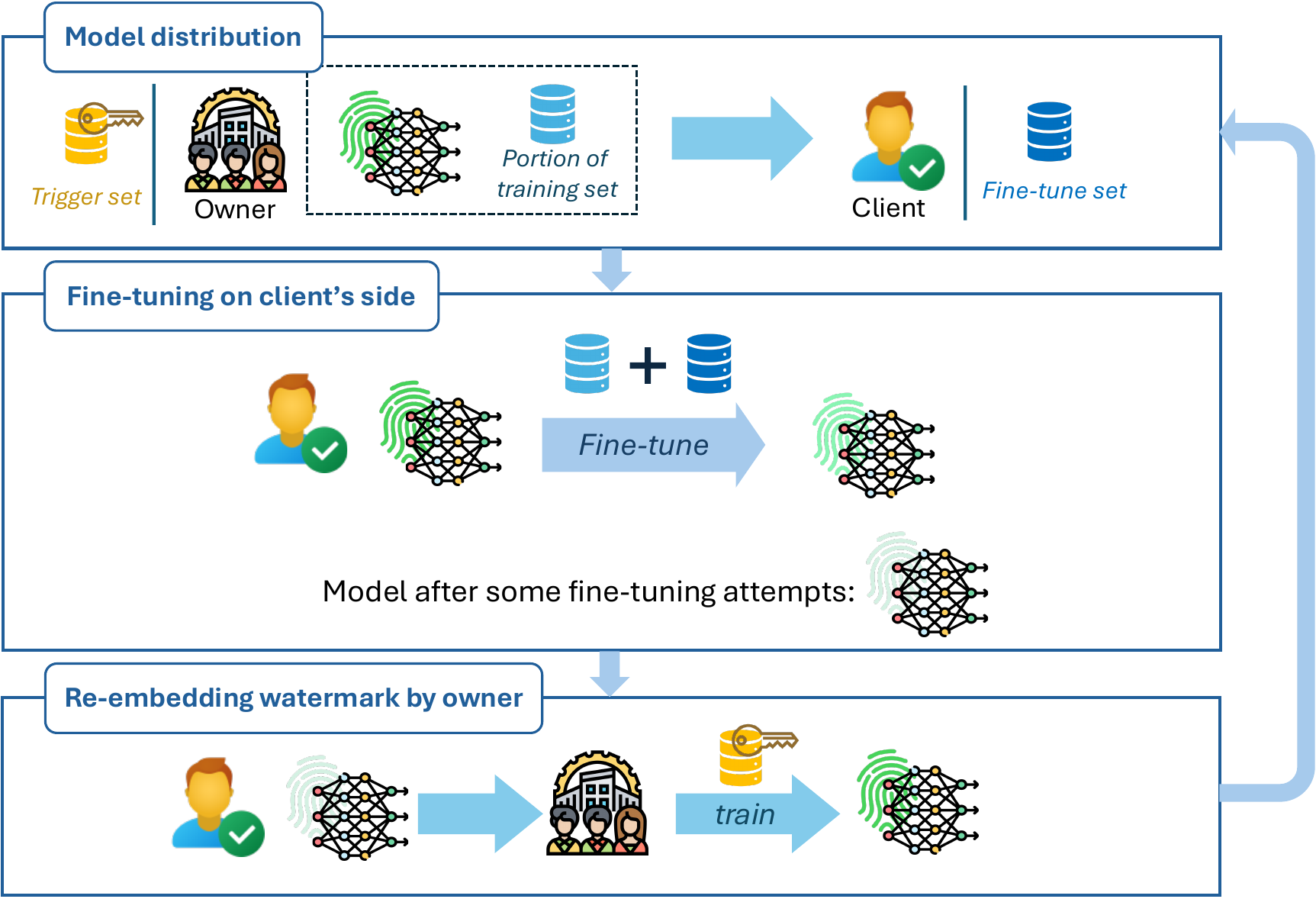}
    \caption{Model fine-tuning with blending between owner's partial training data and client's fine-tuning data. This method helps keep trigger set and fine-tuning set from being disclosed to user and client respectively. Furthermore, the mixture of partial training data helps alleviate watermark vanishing during fine-tuning.}
    \label{fig:threatmodel_mix}
\end{figure}

\textbf{Threat model.} (1) In the scenario that model owner wishes to claim ownership after model is stolen by an adversary: Regarding watermark restoration, one possible question is \textit{why model owner has to bother with original training data, instead of simply retraining the model with trigger samples?} To answer this question, we can think of a scenario where an adversary stole the model, fine-tuned it with custom dataset and deployed it as a public service. To simplify the problem, we assume the adversary's goal for the classification problem is similar to the original task so that they only have to fine-tune the model without customizing the model layers. The model owner, when accessing the suspected model via API access, can try retraining that model with original training data and conduct black-box watermark verification by evaluating the model performance on their secret trigger data, assuming that the cloud service hosting the model supports fine-tune-as-a-service. Another feasible scenario is that an impartial is involved in the verification process to perform the retraining. This retraining process ensures a fair evaluation because trigger accuracy only improves in a previously marked model. This scenario is illustrated in Figure~\ref{fig:threatmodel_retrain}.

(2) In the scenario that there exists a protocol allowing authorized clients to use, fine-tune the model while retaining watermark footprint: With respect to blending original training data during fine-tuning, it helps keep the trigger data from being disclosed to other (even authorized) parties than the model owner while offering enhanced privacy to the client's proprietary fine-tuning dataset. Specifically, we can think of a protocol that allows model owner to securely distribute the watermarked model and share a portion of their training data to an authorized client to mix with their own data for fine-tuning. Here, privacy techniques to ensure a secured storage of model and data in the user's side need to be applied. We assume that the fine-tuning pipeline is predefined by the model owner which automatically blends partial training data with fine-tuning data. After a specific amount of time, e.g. several days/weeks, the client is required to send the model back via that protocol so that model owner can retrain it with trigger data to enhance the watermark before re-distributing it to the client. The idea of such protocol is summarized in Figure~\ref{fig:threatmodel_mix}.

\textbf{Catastrophic forgetting and watermark removal.} In the context of continual learning, when a network is trained on a new task $B$ after old task $A$, the knowledge it has learned from task $A$ gets disrupted. This phenomenon, known as catastrophic forgetting, was first demonstrated in simple multi-layer perceptrons (MLPs)~\cite{mccloskeyCatastrophicInterferenceConnectionist1989}. Since that work, there have been some research attempts to alleviate its effect in deep neural nets, such as~\cite{srivastavaCompeteCompute2013,kirkpatrickOvercomingCatastrophicForgetting2017,coopEnsembleLearningFixed2013}. Kemker et al.~\cite{kemkerMeasuringCatastrophicForgetting2018} developed comprehensive benchmarks for various techniques for forgetting mitigation and concluded that this phenomenon occurs in all common techniques, but at different levels. Regarding the case of backdoor-based watermarking, the classification task on trigger set $\mathcal{D}_{\text{WM}}$ and fine-tuning set $\mathcal{D}_\textsc{Finetune}$ can be viewed as tasks $A$ and $B$ respectively. Since the trigger images are manipulated both in terms of contents and labels, the two tasks can be considered dissimilar. Catastrophic forgetting happens when the watermarked neural network is fine-tuned with $\mathcal{D}_\textsc{Finetune}$.

\subsection{Local Minima and Watermark Restoration}
\textbf{Retraining with original training data.} To achieve the objective as Eq.~\ref{eq:optimization}, model parameters $\theta$ must converge to the local minima that allow $f_\theta$ to give good performance on both $\mathcal{D}_\textsc{Train}$ and $\mathcal{D}_{\text{WM}}$. During fine-tuning on $\mathcal{D}_\textsc{Finetune}$, the parameters shift to locations farther away from the previous locations in parameter space. The new minima are expected to still result in good classification outcomes in $\mathcal{D}_\textsc{Train}$. Nonetheless, model will suffer to maintain a good trigger accuracy on $\mathcal{D}_{\text{WM}}$. It is because $\mathcal{D}_\textsc{Finetune}$ and $\mathcal{D}_\textsc{Train}$ are from the same domain as mentioned in the assumption in Section~\ref{section:preliminaries}, whereas $\mathcal{D}_{\text{WM}}$ is often sampled from a very different domain or probability distribution. How far $\theta$ shift during fine-tuning depends on a variety of aspects, e.g., learning rate, weight decay, optimizer type, difference level between $\mathcal{D}_{\text{WM}}$ and $\mathcal{D}_\textsc{Train}$, etc,. From loss surface geometry viewpoint, if a fine-tuning attack does not completely move the parameters out of the basins of attraction, there is a high chance that the model owner can still pull the parameters back towards the previous local minima again, without retraining the model with trigger data. This can be done simply by introducing original $\mathcal{D}_\textsc{Train}$ in the retraining phase. The intuition behind this is when $\theta$ are still trapped in the previous basins optimized for $\mathcal{D}_\textsc{Train} \cup \mathcal{D}_{\text{WM}}$, further retraining $f_\theta$ solely on $\mathcal{D}_\textsc{Train}$ allows $\theta$ to follow the steepest descent to converge towards these local minima, given appropriate conditions and hyper-parameters. As illustrated in Figure~\ref{fig:threatmodel_retrain}, this idea is useful for a model owner to claim ownership of a suspicious model after it is fine-tuned by an adversary.

\textbf{Fine-tuning with less watermark degradation.} Following the intriguing property of local minima, we propose a technique that alleviates watermark removal from fine-tuning. This is a useful concept, especially in scenarios where model owner distributes their marked model to authorized clients but still allows the clients to fine-tune the model further with their own data to suit their needs as well as offers more privacy to the owner's trigger set and the client's proprietary fine-tuning data, which is shown in Figure~\ref{fig:threatmodel_mix} in previous subsection. The method is, in each iteration, we mix a batch of fine-tuning samples $x_\textsc{Finetune}$ with a batch of training samples $x_\textsc{Train}$. This ensures a balance between incorporating new knowledge and retaining the watermark without further exposing $\mathcal{D}_{\text{WM}}$ to $f_\theta$. Our fine-tuning strategy is detailed in Supplementary Information Algorithm 1.

\begin{figure}[t]
    \centering
    

    \begin{subfigure}[t]{0.15\linewidth}
        \centering
        \caption*{Original}
        \includegraphics[scale=1.2]{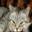}   
    \end{subfigure}
    \begin{subfigure}[t]{0.15\linewidth}
        \centering
        \caption*{Content}
        \includegraphics[scale=1.2]{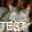}   
    \end{subfigure}
    \begin{subfigure}[t]{0.15\linewidth}
        \centering
        \caption*{Noise}
        \includegraphics[scale=1.2]{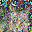}    
    \end{subfigure}
    \begin{subfigure}[t]{0.15\linewidth}
        \centering
        \caption*{FGSM}
        \includegraphics[scale=1.2]{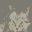} 
    \end{subfigure}
    \begin{subfigure}[t]{0.15\linewidth}
        \centering
        \caption*{Unrelated}
        \includegraphics[scale=1.2]{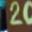} 
    \end{subfigure}
    \caption{Types of trigger images}
    \label{fig:trigger}
\end{figure}

\begin{figure*}[h]
\centering
\\
\ref{resnet_finetune}
\caption{\textit{Trigger accuracy during fine-tuning (ResNet}}
\label{fig:resnet_finetune}
\end{figure*}

\begin{figure*}[h]
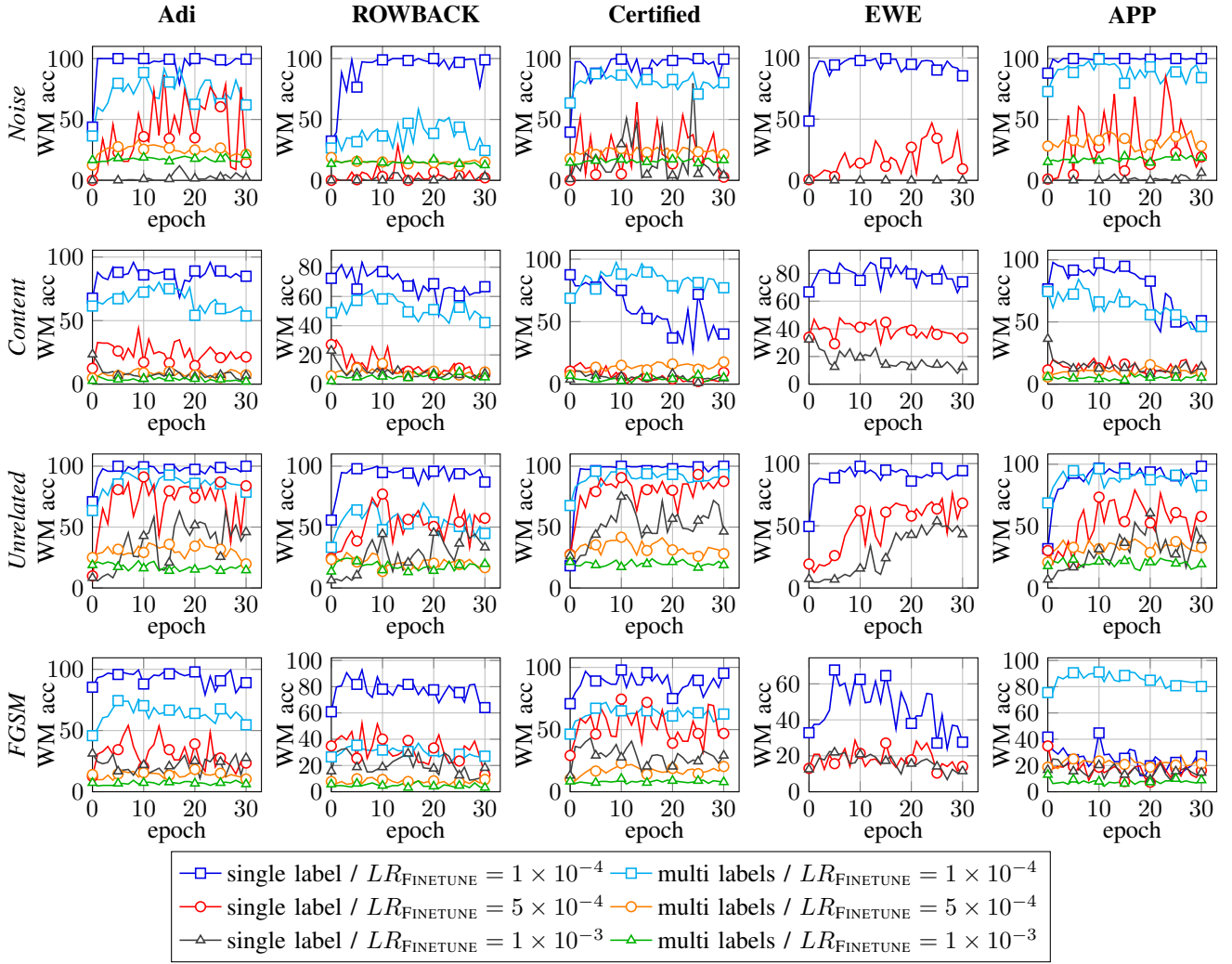

\centering
%
\\
\ref{resnet_retrain}
\caption{\textit{Trigger accuracy during retraining (ResNet)}}
\label{fig:resnet_retrain}
\end{figure*}

\begin{figure*}[!htbp]
    \centering
    \begin{tabular}{@{}m{0.4cm}
    @{}m{\dimexpr0.2\textwidth-0.2cm\relax}
    @{}m{\dimexpr0.2\textwidth-0.2cm\relax}
    @{}m{\dimexpr0.2\textwidth-0.2cm\relax}
    @{}m{\dimexpr0.2\textwidth-0.2cm\relax}
    @{}m{\dimexpr0.2\textwidth-0.2cm\relax}
    @{}}
    \raisebox{-\baselineskip}{\rotatebox{90}{\textit{Noise}}}
    & \begin{subfigure}[b]{\linewidth}
        \centering
        \caption*{\textbf{Adi}}
        \includegraphics[width=0.75\linewidth]{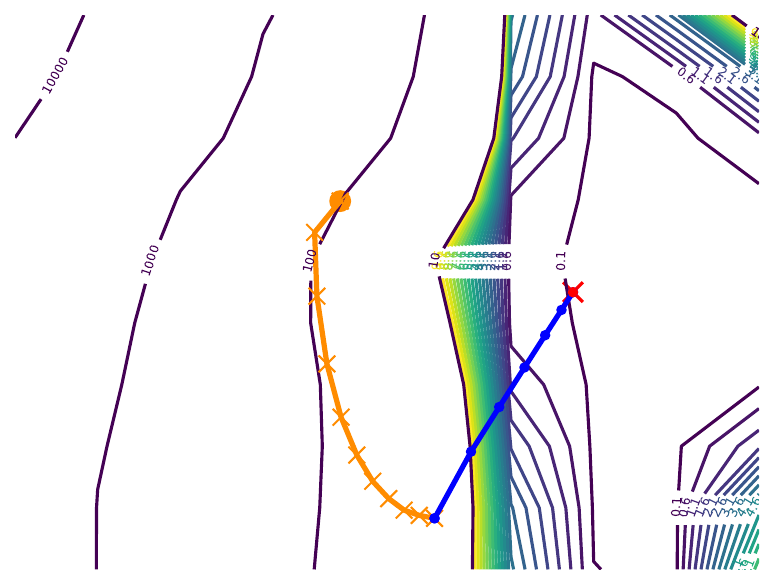} 
    \end{subfigure}  
    & \begin{subfigure}[b]{\linewidth}
        \centering
        \caption*{\textbf{ROWBACK}}
        \includegraphics[width=0.75\linewidth]{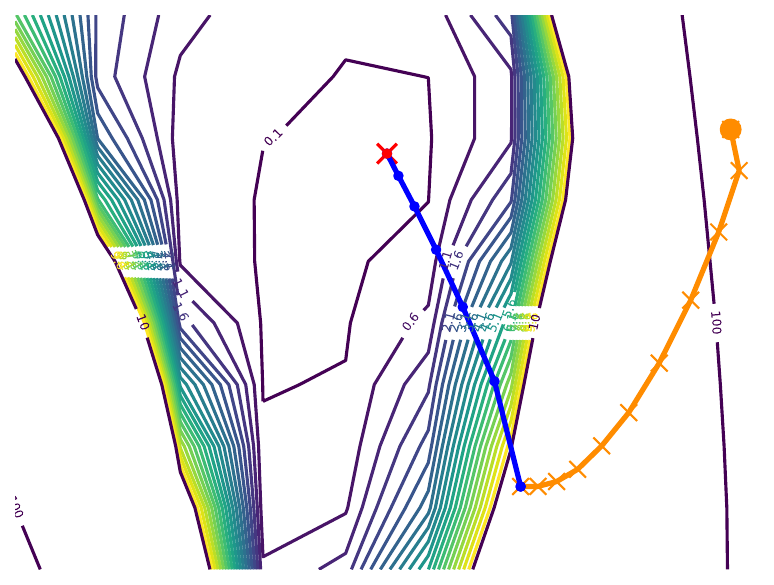} 
    \end{subfigure}
    & \begin{subfigure}[b]{\linewidth}
        \centering
        \caption*{\textbf{Certified}}
        \includegraphics[width=0.75\linewidth]{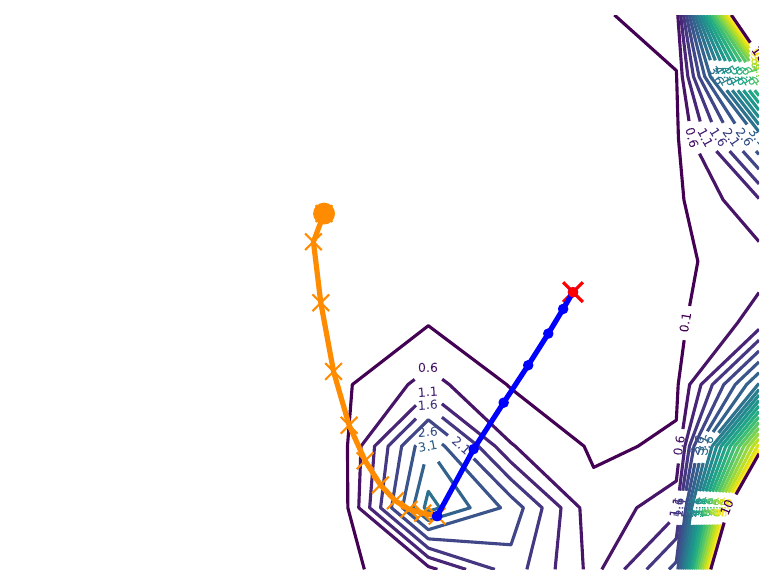} 
        \end{subfigure}  
    & \begin{subfigure}[b]{\linewidth}
        \centering
        \caption*{\textbf{EWE}}
        \includegraphics[width=0.75\linewidth]{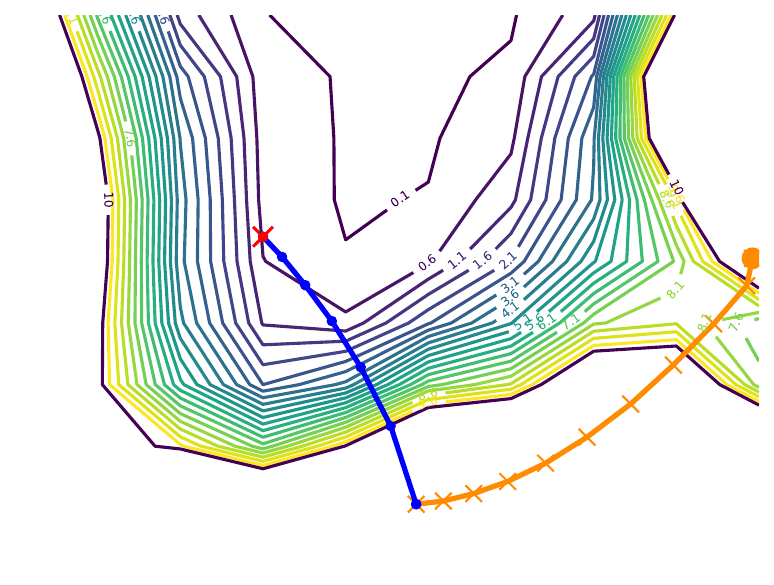} 
    \end{subfigure}
    & \begin{subfigure}[b]{\linewidth}
        \centering
        \caption*{\textbf{APP}}
        \includegraphics[width=0.75\linewidth]{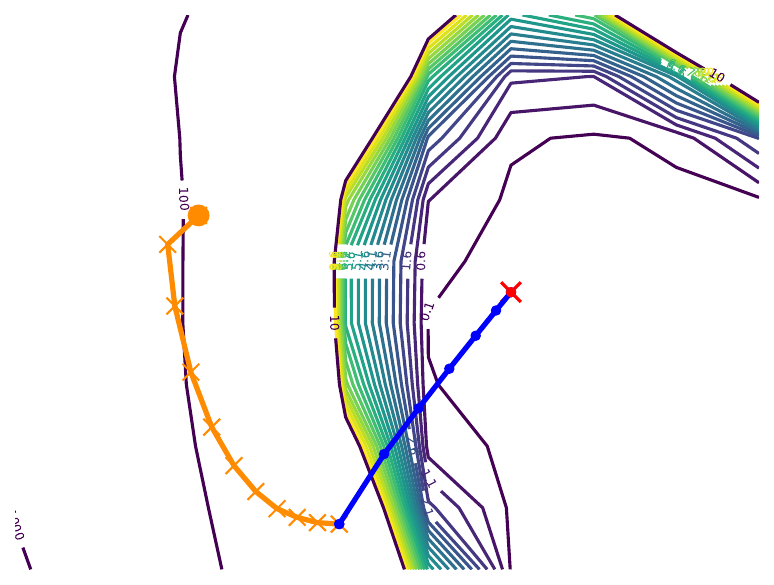} 
    \end{subfigure} \\
\end{tabular}
    \caption{\textit{Loss landscape visualization for fine-tuning attack (ResNet)} - The contours illustrate trigger loss, \textcolor{orange}{\textit{orange}} lines are the fine-tuning phase while \textcolor{blue}{\textit{blue}} lines represent retraining. It is observable that retraining helps steering the trajectory back to near the local minima. Note: for these visualizations, the projection of model checkpoints near the end of retraining are more accurate than earlier checkpoints. Therefore, we only visualize checkpoints started from fine-tuning stage.}
    \label{fig:contours}
\end{figure*}

\section{Experiments and Results}\label{experiment}

We conducted extensive experiments, first to evaluate the resistance of five watermark schemes Adi et al.~\cite{adiTurningYourWeakness2018}, ROWBACK~\cite{chattopadhyayROWBACKRObustWatermarking2021}, certified watermark~\cite{bansalCertifiedNeuralNetwork2022}, entangled watermark (EWE)~\cite{jiaEntangledWatermarksDefense2021} and adversarial parametric perturbation (APP)~\cite{ganRobustModelWatermark2023} to fine-tuning attack. We choose these schemes to include in our experiments because they well reflect the advancement of backdoor-based watermarks, from the first to one of the most recent schemes. However, we could not include too many schemes due to limited computing resources and implementation capability. Moreover, after putting effort into implementation, these are the schemes that we could have access to their original code and achieve desirable results as proposed in their papers.

In the second experiment, we investigate whether the watermark can be restored by retraining the neural networks solely with the original training data from the first training phase. We found that the original training data, interestingly, help bring the watermark back without the presence of trigger data in retraining phase, depending on trigger set type. This intriguing result has led us to the third experiment, in which the original training data are mixed with fine-tuning data, to mitigate the erosion of watermarks.

Our experiments can be reproduced with code available on GitHub\footnote{Code: \url{https://github.com/anhtu96/dnn-watermark-persistence}}.

\subsection{Experimental Design}
We use CIFAR-10 dataset~\cite{Krizhevsky09learningmultiple} for the main classification task, which consists of 50K training images across 10 object classes. From the original training dataset, 200 samples are randomly chosen to create trigger set, whose generation process is detailed in the later part. Regarding unrelated trigger set, we sample 200 images from the Street View House Numbers (SVHN) dataset~\cite{netzer2011reading}. In the remaining subset of the original training set, $70\%$ are used for model pretraining and the remaining $30\%$ are used for fine-tuning. We denote the subsets for pretraining, watermark embedding and fine-tuning $\mathcal{D}_\textsc{Train}, \mathcal{D}_{\text{WM}}$ and $\mathcal{D}_\textsc{Finetune}$ respectively. Most of the experiments are run on our lab machine with a single NVIDIA RTX A4000 GPU, while some parts of them are run on Singapore's NSCC ASPIRE 2A cluster with NVIDIA A100 GPU.

\textbf{Trigger samples generation.} There are many ways to generate trigger set. For example, Zhang et al.~\cite{zhangProtectingIntellectualProperty2018} proposed three techniques such as adding random noise, embedding contents (logo, text, etc,.) or using out-of-distribution samples for trigger data. In this study, we experiment with all these methods as well as the one proposed by Chattopadhyay \& Chattopadhyay~\cite{chattopadhyayROWBACKRObustWatermarking2021}, where trigger samples are adversarial images crafted from clean training samples using fast gradient sign methond (FGSM)~\cite{goodfellowExplainingHarnessingAdversarial2015}. Figure~\ref{fig:trigger} visualizes examples of trigger types for class \textit{cat}.

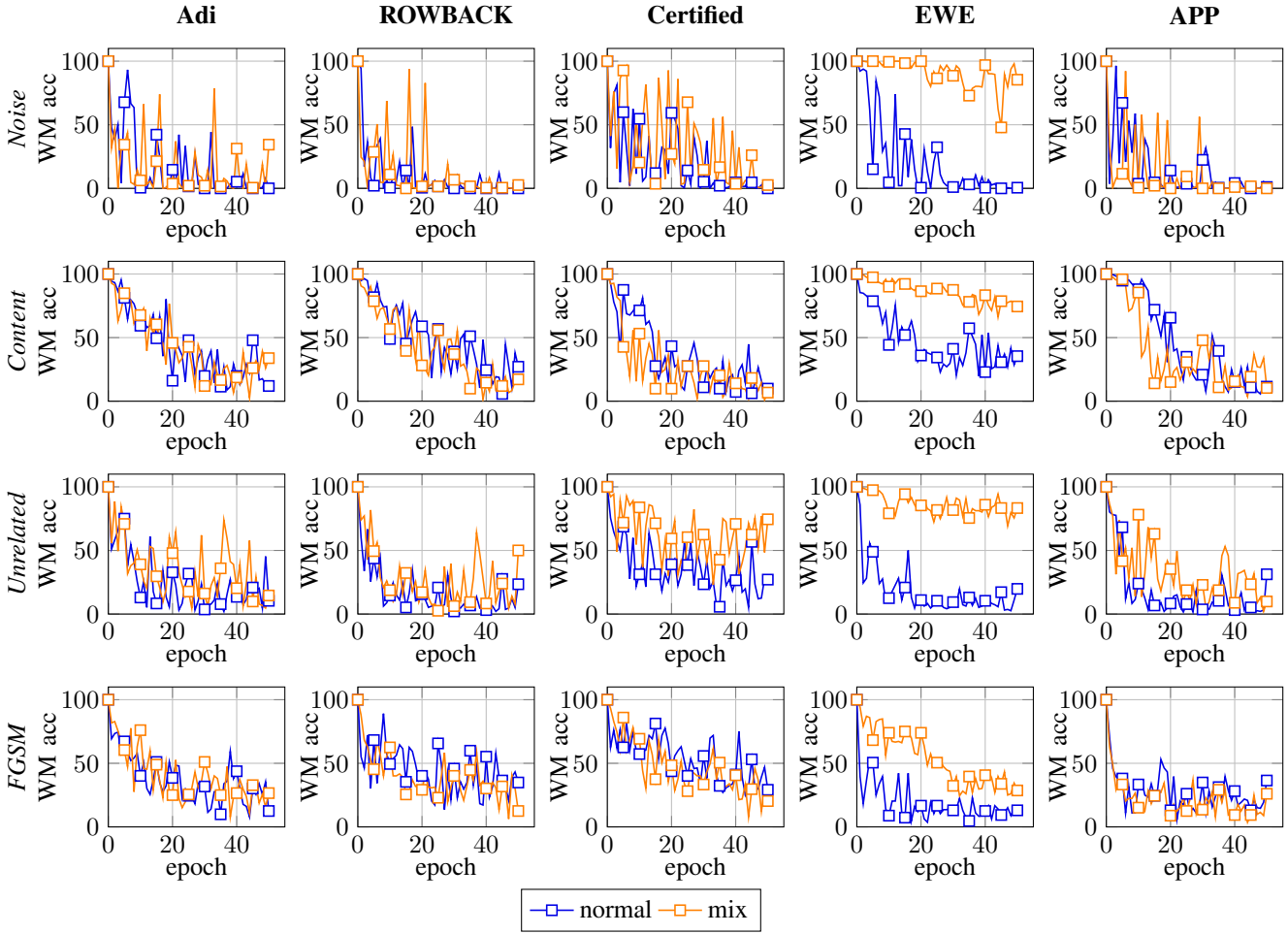
\begin{figure*}[h]
\centering
\begin{tikzpicture}
    \begin{groupplot}[group style={group size=5 by 4},height=3.5cm,width=4cm]
    \nextgroupplot[
        legend to name=resnet_finetune_mix,
        legend entries={normal, mix},
        legend columns=2,
        title={\textbf{Adi}},
        grid=major,
        ylabel style={align=center},
        ylabel shift=-5pt,
        ylabel={\textit{Noise}\\WM acc},
        xlabel shift=-5pt, xlabel={epoch},
        ymin=0,
        xmin=0
    ]
        \addplot[DarkBlue, every mark/.append style={solid, fill=white},mark=square*,mark repeat=5,mark size=2pt, line width=0.6pt] table[x=epoch,y={adi_noise_resnet_cifar10_medlr},col sep=comma]{"csv/wandb_ftal_finetune_acc.csv"};
        \addplot[orange,every mark/.append style={solid, fill=white},mark=square*,mark repeat=5,mark size=2pt, line width=0.6pt] table[x=epoch,y={adi_noise_resnet_cifar10_medlr},col sep=comma]{"csv/wandb_ftal_mix_acc.csv"};
    \nextgroupplot[
        title={\textbf{ROWBACK}},
        grid=major,
        ylabel shift=-10pt, ylabel={WM acc},
        xlabel shift=-5pt, xlabel={epoch},
        ymin=0,
        xmin=0
    ]
        \addplot[DarkBlue, every mark/.append style={solid, fill=white},mark=square*,mark repeat=5,mark size=2pt, line width=0.6pt] table[x=epoch,y={rowback_noise_resnet_cifar10_medlr},col sep=comma]{"csv/wandb_ftal_finetune_acc.csv"};
        \addplot[orange,every mark/.append style={solid, fill=white},mark=square*,mark repeat=5,mark size=2pt, line width=0.6pt] table[x=epoch,y={rowback_noise_resnet_cifar10_medlr},col sep=comma]{"csv/wandb_ftal_mix_acc.csv"};
     \nextgroupplot[
        title={\textbf{Certified}},
        grid=major,
        ymin=0,
        xmin=0,
        ylabel shift=-10pt, ylabel={WM acc},
        xlabel shift=-5pt, xlabel={epoch},
     ]
        \addplot[DarkBlue, every mark/.append style={solid, fill=white},mark=square*,mark repeat=5,mark size=2pt, line width=0.6pt] table[x=epoch,y={certified_noise_resnet_cifar10_medlr},col sep=comma]{"csv/wandb_ftal_finetune_acc.csv"};
        \addplot[orange,every mark/.append style={solid, fill=white},mark=square*,mark repeat=5,mark size=2pt, line width=0.6pt] table[x=epoch,y={certified_noise_resnet_cifar10_medlr},col sep=comma]{"csv/wandb_ftal_mix_acc.csv"};
     \nextgroupplot[
        title={\textbf{EWE}},
        grid=major,
        ymin=0,
        xmin=0,
        ylabel shift=-10pt, ylabel={WM acc},
        xlabel shift=-5pt, xlabel={epoch},
     ]
        \addplot[DarkBlue, every mark/.append style={solid, fill=white},mark=square*,mark repeat=5,mark size=2pt, line width=0.6pt] table[x=epoch,y={ewe_noise_resnet_cifar10_medlr},col sep=comma]{"csv/wandb_ftal_finetune_acc.csv"};
        \addplot[orange,every mark/.append style={solid, fill=white},mark=square*,mark repeat=5,mark size=2pt, line width=0.6pt] table[x=epoch,y={ewe_noise_resnet_cifar10_medlr},col sep=comma]{"csv/wandb_ftal_mix_acc.csv"};
    \nextgroupplot[
        title={\textbf{APP}},
        grid=major,
        ylabel style={align=center},
        ylabel shift=-10pt, ylabel={WM acc},
        xlabel shift=-5pt, xlabel={epoch},
        ymin=0,
        xmin=0
    ]
        \addplot[DarkBlue, every mark/.append style={solid, fill=white},mark=square*,mark repeat=5,mark size=2pt, line width=0.6pt] table[x=epoch,y={app_noise_resnet_cifar10_medlr},col sep=comma]{"csv/wandb_ftal_finetune_acc.csv"};
        \addplot[orange,every mark/.append style={solid, fill=white},mark=square*,mark repeat=5,mark size=2pt, line width=0.6pt] table[x=epoch,y={app_noise_resnet_cifar10_medlr},col sep=comma]{"csv/wandb_ftal_mix_acc.csv"};
    \nextgroupplot[
        grid=major,
        ylabel style={align=center},
        ylabel shift=-5pt,
        ylabel={\textit{Content}\\WM acc},
        xlabel shift=-5pt, xlabel={epoch},
        ymin=0,
        xmin=0
    ]
        \addplot[DarkBlue, every mark/.append style={solid, fill=white},mark=square*,mark repeat=5,mark size=2pt, line width=0.6pt] table[x=epoch,y={adi_textoverlay_resnet_cifar10_medlr},col sep=comma]{"csv/wandb_ftal_finetune_acc.csv"};
        \addplot[orange,every mark/.append style={solid, fill=white},mark=square*,mark repeat=5,mark size=2pt, line width=0.6pt] table[x=epoch,y={adi_textoverlay_resnet_cifar10_medlr},col sep=comma]{"csv/wandb_ftal_mix_acc.csv"};
     \nextgroupplot[
        grid=major,
        ylabel style={align=center},
        ylabel shift=-10pt,
        ylabel={WM acc},
        xlabel shift=-5pt, xlabel={epoch},
        ymin=0,
        xmin=0
     ]
        \addplot[DarkBlue, every mark/.append style={solid, fill=white},mark=square*,mark repeat=5,mark size=2pt, line width=0.6pt] table[x=epoch,y={rowback_textoverlay_resnet_cifar10_medlr},col sep=comma]{"csv/wandb_ftal_finetune_acc.csv"};
        \addplot[orange,every mark/.append style={solid, fill=white},mark=square*,mark repeat=5,mark size=2pt, line width=0.6pt] table[x=epoch,y={rowback_textoverlay_resnet_cifar10_medlr},col sep=comma]{"csv/wandb_ftal_mix_acc.csv"};
     \nextgroupplot[
        grid=major,
        ylabel style={align=center},
        ylabel shift=-10pt,
        ylabel={WM acc},
        xlabel shift=-5pt, xlabel={epoch},
        ymin=0,
        xmin=0
     ]
        \addplot[DarkBlue, every mark/.append style={solid, fill=white},mark=square*,mark repeat=5,mark size=2pt, line width=0.6pt] table[x=epoch,y={certified_textoverlay_resnet_cifar10_medlr},col sep=comma]{"csv/wandb_ftal_finetune_acc.csv"};
        \addplot[orange,every mark/.append style={solid, fill=white},mark=square*,mark repeat=5,mark size=2pt, line width=0.6pt] table[x=epoch,y={certified_textoverlay_resnet_cifar10_medlr},col sep=comma]{"csv/wandb_ftal_mix_acc.csv"};
        
    \nextgroupplot[
        grid=major,
        ylabel style={align=center},
        ylabel shift=-10pt,
        ylabel={WM acc},
        xlabel shift=-5pt, xlabel={epoch},
        ymin=0,
        xmin=0
    ]
        \addplot[DarkBlue, every mark/.append style={solid, fill=white},mark=square*,mark repeat=5,mark size=2pt, line width=0.6pt] table[x=epoch,y={ewe_textoverlay_resnet_cifar10_medlr},col sep=comma]{"csv/wandb_ftal_finetune_acc.csv"};
        \addplot[orange,every mark/.append style={solid, fill=white},mark=square*,mark repeat=5,mark size=2pt, line width=0.6pt] table[x=epoch,y={ewe_textoverlay_resnet_cifar10_medlr},col sep=comma]{"csv/wandb_ftal_mix_acc.csv"};
    \nextgroupplot[
        grid=major,
        ylabel style={align=center},
        ylabel shift=-10pt,
        ylabel={WM acc},
        xlabel shift=-5pt, xlabel={epoch},
        ymin=0,
        xmin=0
    ]
        \addplot[DarkBlue, every mark/.append style={solid, fill=white},mark=square*,mark repeat=5,mark size=2pt, line width=0.6pt] table[x=epoch,y={app_textoverlay_resnet_cifar10_medlr},col sep=comma]{"csv/wandb_ftal_finetune_acc.csv"};
        \addplot[orange,every mark/.append style={solid, fill=white},mark=square*,mark repeat=5,mark size=2pt, line width=0.6pt] table[x=epoch,y={app_textoverlay_resnet_cifar10_medlr},col sep=comma]{"csv/wandb_ftal_mix_acc.csv"};
    \nextgroupplot[
        grid=major,
        ylabel style={align=center},
        ylabel shift=-5pt,
        ylabel={\textit{Unrelated}\\WM acc},
        xlabel shift=-5pt, xlabel={epoch},
        ymin=0,
        xmin=0
    ]
        \addplot[DarkBlue, every mark/.append style={solid, fill=white},mark=square*,mark repeat=5,mark size=2pt, line width=0.6pt] table[x=epoch,y={adi_unrelated_resnet_cifar10_medlr},col sep=comma]{"csv/wandb_ftal_finetune_acc.csv"};
        \addplot[orange,every mark/.append style={solid, fill=white},mark=square*,mark repeat=5,mark size=2pt, line width=0.6pt] table[x=epoch,y={adi_unrelated_resnet_cifar10_medlr},col sep=comma]{"csv/wandb_ftal_mix_acc.csv"};
     \nextgroupplot[
        grid=major,
        ylabel style={align=center},
        ylabel shift=-10pt,
        ylabel={WM acc},
        xlabel shift=-5pt, xlabel={epoch},
        ymin=0,
        xmin=0
     ]
        \addplot[DarkBlue, every mark/.append style={solid, fill=white},mark=square*,mark repeat=5,mark size=2pt, line width=0.6pt] table[x=epoch,y={rowback_unrelated_resnet_cifar10_medlr},col sep=comma]{"csv/wandb_ftal_finetune_acc.csv"};
        \addplot[orange,every mark/.append style={solid, fill=white},mark=square*,mark repeat=5,mark size=2pt, line width=0.6pt] table[x=epoch,y={rowback_unrelated_resnet_cifar10_medlr},col sep=comma]{"csv/wandb_ftal_mix_acc.csv"};
     \nextgroupplot[
        grid=major,
        ylabel style={align=center},
        ylabel shift=-10pt,
        ylabel={WM acc},
        xlabel shift=-5pt, xlabel={epoch},
        ymin=0,
        xmin=0
     ]
        \addplot[DarkBlue, every mark/.append style={solid, fill=white},mark=square*,mark repeat=5,mark size=2pt, line width=0.6pt] table[x=epoch,y={certified_unrelated_resnet_cifar10_medlr},col sep=comma]{"csv/wandb_ftal_finetune_acc.csv"};
        \addplot[orange,every mark/.append style={solid, fill=white},mark=square*,mark repeat=5,mark size=2pt, line width=0.6pt] table[x=epoch,y={certified_unrelated_resnet_cifar10_medlr},col sep=comma]{"csv/wandb_ftal_mix_acc.csv"};
        
    \nextgroupplot[
        grid=major,
        ylabel style={align=center},
        ylabel shift=-10pt,
        ylabel={WM acc},
        xlabel shift=-5pt, xlabel={epoch},
        ymin=0,
        xmin=0
    ]
        \addplot[DarkBlue, every mark/.append style={solid, fill=white},mark=square*,mark repeat=5,mark size=2pt, line width=0.6pt] table[x=epoch,y={ewe_unrelated_resnet_cifar10_medlr},col sep=comma]{"csv/wandb_ftal_finetune_acc.csv"};
        \addplot[orange,every mark/.append style={solid, fill=white},mark=square*,mark repeat=5,mark size=2pt, line width=0.6pt] table[x=epoch,y={ewe_unrelated_resnet_cifar10_medlr},col sep=comma]{"csv/wandb_ftal_mix_acc.csv"};
    \nextgroupplot[
        grid=major,
        ylabel style={align=center},
        ylabel shift=-10pt,
        ylabel={WM acc},
        xlabel shift=-5pt, xlabel={epoch},
        ymin=0,
        xmin=0
    ]
        \addplot[DarkBlue, every mark/.append style={solid, fill=white},mark=square*,mark repeat=5,mark size=2pt, line width=0.6pt] table[x=epoch,y={app_unrelated_resnet_cifar10_medlr},col sep=comma]{"csv/wandb_ftal_finetune_acc.csv"};
        \addplot[orange,every mark/.append style={solid, fill=white},mark=square*,mark repeat=5,mark size=2pt, line width=0.6pt] table[x=epoch,y={app_unrelated_resnet_cifar10_medlr},col sep=comma]{"csv/wandb_ftal_mix_acc.csv"};

    \nextgroupplot[
        grid=major,
        ylabel style={align=center},
        ylabel shift=-5pt,
        ylabel={\textit{FGSM}\\WM acc},
        xlabel shift=-5pt, xlabel={epoch},
        ymin=0,
        xmin=0
    ]
        \addplot[DarkBlue, every mark/.append style={solid, fill=white},mark=square*,mark repeat=5,mark size=2pt, line width=0.6pt] table[x=epoch,y={adi_adv_resnet_cifar10_medlr},col sep=comma]{"csv/wandb_ftal_finetune_acc.csv"};
        \addplot[orange,every mark/.append style={solid, fill=white},mark=square*,mark repeat=5,mark size=2pt, line width=0.6pt] table[x=epoch,y={adi_adv_resnet_cifar10_medlr},col sep=comma]{"csv/wandb_ftal_mix_acc.csv"};
     \nextgroupplot[
        grid=major,
        ylabel style={align=center},
        ylabel shift=-10pt,
        ylabel={WM acc},
        xlabel shift=-5pt, xlabel={epoch},
        ymin=0,
        xmin=0
     ]
        \addplot[DarkBlue, every mark/.append style={solid, fill=white},mark=square*,mark repeat=5,mark size=2pt, line width=0.6pt] table[x=epoch,y={rowback_adv_resnet_cifar10_medlr},col sep=comma]{"csv/wandb_ftal_finetune_acc.csv"};
        \addplot[orange,every mark/.append style={solid, fill=white},mark=square*,mark repeat=5,mark size=2pt, line width=0.6pt] table[x=epoch,y={rowback_adv_resnet_cifar10_medlr},col sep=comma]{"csv/wandb_ftal_mix_acc.csv"};
     \nextgroupplot[
        grid=major,
        ylabel style={align=center},
        ylabel shift=-10pt,
        ylabel={WM acc},
        xlabel shift=-5pt, xlabel={epoch},
        ymin=0,
        xmin=0
     ]
        \addplot[DarkBlue, every mark/.append style={solid, fill=white},mark=square*,mark repeat=5,mark size=2pt, line width=0.6pt] table[x=epoch,y={certified_adv_resnet_cifar10_medlr},col sep=comma]{"csv/wandb_ftal_finetune_acc.csv"};
        \addplot[orange,every mark/.append style={solid, fill=white},mark=square*,mark repeat=5,mark size=2pt, line width=0.6pt] table[x=epoch,y={certified_adv_resnet_cifar10_medlr},col sep=comma]{"csv/wandb_ftal_mix_acc.csv"};
        
    \nextgroupplot[
        grid=major,
        ylabel style={align=center},
        ylabel shift=-10pt,
        ylabel={WM acc},
        xlabel shift=-5pt, xlabel={epoch},
        ymin=0,
        xmin=0
    ]
        \addplot[DarkBlue, every mark/.append style={solid, fill=white},mark=square*,mark repeat=5,mark size=2pt, line width=0.6pt] table[x=epoch,y={ewe_adv_resnet_cifar10_medlr},col sep=comma]{"csv/wandb_ftal_finetune_acc.csv"};
        \addplot[orange,every mark/.append style={solid, fill=white},mark=square*,mark repeat=5,mark size=2pt, line width=0.6pt] table[x=epoch,y={ewe_adv_resnet_cifar10_medlr},col sep=comma]{"csv/wandb_ftal_mix_acc.csv"};
    \nextgroupplot[
        grid=major,
        ylabel style={align=center},
        ylabel shift=-10pt,
        ylabel={WM acc},
        xlabel shift=-5pt, xlabel={epoch},
        ymin=0,
        xmin=0
    ]
        \addplot[DarkBlue, every mark/.append style={solid, fill=white},mark=square*,mark repeat=5,mark size=2pt, line width=0.6pt] table[x=epoch,y={app_adv_resnet_cifar10_medlr},col sep=comma]{"csv/wandb_ftal_finetune_acc.csv"};
        \addplot[orange,every mark/.append style={solid, fill=white},mark=square*,mark repeat=5,mark size=2pt, line width=0.6pt] table[x=epoch,y={app_adv_resnet_cifar10_medlr},col sep=comma]{"csv/wandb_ftal_mix_acc.csv"};
    \end{groupplot}
\end{tikzpicture}\\
\ref{resnet_finetune_mix}
\caption{\textit{Comparison of trigger accuracies between mixing and without mixing of training data $\mathcal{D}_\textsc{Train}$} (ResNet)}
\label{fig:resnet_finetune_mix}
\end{figure*}

\textbf{Trigger data labeling schemes.} For each trigger set type, we use two different labeling schemes to evaluate model performance. In the first scheme, we assign a fixed label to all 200 trigger samples, here we assign label \textit{airplane} to all images. For the second scheme, we assign various labels to trigger samples. Details on how we implement this scheme are described in Supplementary Information Algorithm 2.

\textbf{NN models and training procedure.} We experiment with ResNet-18~\cite{heDeepResidualLearning2016} and ViT-S~\cite{dosovitskiyImageWorth16x162020}. In each trial, we first pretrain the model with $\mathcal{D}_\textsc{Train}$ and embed the watermark into it using the trigger set $\mathcal{D}_{\text{WM}}$. During training, $\mathcal{D}_\textsc{Train}$ is poisoned with trigger samples and the model is trained with this mixed set. In terms of training mechanism, five different training techniques are employed to embed the watermark as proposed by Adi et al.~\cite{adiTurningYourWeakness2018}, N. Chattopadhyay \& A. Chattopadhyay~\cite{chattopadhyayROWBACKRObustWatermarking2021}, Bansal et al.~\cite{bansalCertifiedNeuralNetwork2022}, Jia et al.~\cite{jiaEntangledWatermarksDefense2021} and Gan et al.~\cite{ganRobustModelWatermark2023}. We denote these watermark embedding techniques as Adi, ROWBACK, Certified, EWE and APP respectively. For each scheme, we experiment with 4 trigger set types, 2 labeling schemes (fixed label, multiple labels) and 2 NN types (ResNet-18, ViT). However, for ROWBACK and EWE, we do not test them with ViT since their layer customizations were originally implemented for ResNet-based networks. It is worth mentioning that scheme Certified has much longer training time per epoch compared to the others, since its training procedure requires calculating mean gradient of many noised copies of the original parameters. In terms of EWE, our implementation was inspired from Watermark Robustness Toolbox~\cite{lukasSoKHowRobust2022}. The details for all hyper-parameters are mentioned in Supplementary Information Table I.

\textbf{Model performance.} A popular metric to measure the existence of watermark is trigger set accuracy. Supplementary Information Table II illustrates the test accuracy and trigger accuracy of the NN models after pretraining/watermark embedding. It can be seen that regarding ResNet-18, schemes Adi, Certified and APP give comparable performance in terms of test accuracy and trigger accuracy, whereas EWE has lowest test accuracy due to its trade-off between task performance and entanglement, and ROWBACK achieves lowest watermark performance among these five schemes. When it comes to ViT, Adi and APP enjoy similar performance, whereas Certified is slightly behind. In terms of watermark fidelity, we compared the test accuracy with clean models trained using the same hyperparameters, which achieved 87.86\% (ResNet-18) and 77.06\% (ViT) respectively. It is observable that most tested schemes retain the main task accuracy quite well, except ROWBACK and EWE due to the per-layer training mechanism (ROWBACK) and entangled representations between clean and trigger samples (EWE).

\subsection{Empirical Analysis}
We conducted various experiments to assess the watermark persistence of the five schemes. In our first experiment, we simply measure the trigger accuracy after model fine-tuning. Then, we empirically validate the concept of watermark reinstatement by retraining the DNNs with the original training set $\mathcal{D}_\textsc{Train}$. Furthermore, we use loss landscape analysis to investigate the optimization trajectory of model parameters in such scenarios. And finally, our final experiment is about incorporating the original training data into fine-tuning phase to examine its effectiveness in alleviating watermark degradation.

\subsubsection{Fine-tuning}\label{subsection:finetune}
We measure watermark removal level after fine-tuning. This involves training the model with extended data to incorporate broader knowledge into the model. From the adversary's point of view, fine-tuning is equivalent to removal attack by gradually training the model with new data. After NNs are pretrained and watermarked, we fine-tune them with $\mathcal{D}_\textsc{Finetune}$. Then, we evaluate the level of diminishing in trigger accuracy when fine-tuning with different learning rate $1\times 10^{-4}, 5\times 10^{-4}, 1\times 10^{-3}$. It is worth noting that for all of the following figures in this paper, $LR_\textsc{Finetune}$ means learning rate during fine-tuning phase. Figure~\ref{fig:resnet_finetune} and Supplementary Information Figure 1 represent the trends of watermark accuracies when experimenting with ResNet-18 and ViT-S respectively. It can be observed that the watermark accuracies of ROWBACK go down more significantly than the others in most cases. Regarding trigger samples with noise, the watermark accuracies wiggle a lot, even when the models are fine-tuned with small learning rate $1\times 10^{-4}$. In the case of ViT models, the fluctuations in watermark accuracies are milder.

In terms of the effect of labeling schemes to watermark performance, it is not really obvious to distinguish the difference between single-label and multi-label schemes for ResNet-18 models. However, in the majority of cases, the accuracies on trigger samples with fixed label are higher than the accuracies on those with multiple labels. This trend becomes clearer when it comes to fine-tuning ViT models. A possible explanation for this is that when using multiple labels, a greater capacity of NN is needed to distinguish between many ``abnormal'' features. Furthermore, a greater number of classes means fewer trigger samples per class, which might not be enough for the model to learn. For ViT, we can see that APP is able to maintain good watermark performance when fine-tuned at small learning rate.

\subsubsection{Retraining for watermark restoration}\label{subsection:retraining}
Here we do empirical study on restoring watermark after it gets eroded after incremental training / fine-tuning attack, without reintroducing  trigger data into model training. This idea will be useful in real-world scenario described in Section~\ref{section:preliminaries}, which helps mitigate the risk of leaking owner's secret key and client's proprietary data.

After fine-tuning and having the watermark degraded, we retrain the model with the initial training set $\mathcal{D}_\textsc{Train}$. The models are trained for 30 epochs with Adam optimizer. The details for learning rate are listed in Supplementary Information Table I. Figure~\ref{fig:resnet_retrain} and Supplementary Information Figure 2 illustrate how the watermark is reinstated by using solely original training data. For models fine-tuned with small learning rate $1\times 10^{-4}$, retraining helps regain watermark footprint for all watermark schemes in the case of FGSM, noised and unrelated trigger samples regardless of NN types. In terms of noised trigger, the trigger accuracy can be brought back to up to $100\%$ in small learning rate scenarios, but the trend is not very clear with medium and big fine-tune learning rates, especially for ViT models. Regarding FGSM triggers, the watermark can be restored when the NN is fine-tuned with small or medium learning rate, though this upward trend is not observable all the time. 

It can be seen from our experiments that watermark restoration occurs most clearly with unrelated trigger samples. One viable hypothesis for this is since unrelated trigger images do not share (or share very few) common features with training images and fine-tuning images, fine-tuning NN does not significantly mess with the model capacity to classify trigger samples. In terms of ResNet-18 (Figure~\ref{fig:resnet_retrain}), the trigger accuracies increase with respect to all fine-tuning learning rates, whereas, the trigger accuracies oscillate between low and high values for medium and big learning rates. Nevertheless, this is still useful to claim ownership because only previously watermarked model can give high trigger accuracy when retrained, even if this accuracy fluctuates. As for ViT models (Supplementary Information Figure 2), the restoration trend is very obvious with respect to scheme APP, for all fine-tuning learning rates. Meanwhile, the watermark restoration seems to be less noticeable in terms of text-overlaid trigger samples. In some cases, the accuracy even gets worse during retraining. This might be due to the fact that text-overlaid samples share many common features with the original training samples and fine-tuning samples, except for the areas of embedded text. As a result, fine-tuning and retraining mess with the watermark's existence.

\textbf{Loss landscape analysis.} To further explore how retraining with training data affects the watermark, we visualize the loss landscape with learning trajectory. In this study, we use filter normalization method for loss landscape visualization and PCA for optimization trajectory plotting, which was proposed by Li et al.~\cite{liVisualizingLossLandscape2018}. Figure~\ref{fig:contours} shows 2D contours for trigger loss along the approximate learning trajectories of model parameters when fine-tuned with small learning rate $1\times 10^{-4}$, regarding ResNet-18 models. Due to the similarity in watermark performance for different labeling schemes, we only visualize the loss landscapes in the case of single-label trigger data. It is obvious that during the retraining stage (\textcolor{blue}{blue} lines), the learning trajectories turn sharply, with the exception of EWE on FGSM trigger. And in most cases, the parameters move towards the contour lines with smaller loss values. From our inspection, the PCA projection of final point in the trajectory correctly corresponds to the actual trigger loss from the contours. However, the approximations for points in earlier epochs do not completely match their actual trigger loss values. In other words, if a projected point is closer to the final projected point in parameter space, its approximation is more accurate than further points. Nonetheless, the PCA approximation is still very beneficial to study about the optimization trend.

\subsubsection{Fine-tuning with mixed data}
The intriguing phenomenon of watermark reappearance has led us to a follow-up question - \textit{Can introducing training samples into fine-tuning help alleviate watermark vanishing?} To test this out, we modify the fine-tuning process as described in Supplementary Information Algorithm 1, where during fine-tuning, a random batch of training data $\mathcal{D}_\textsc{Train}$ is fed into the training loop after a specific number of epochs. Intuitively, mixing a portion of the training data helps guide the parameters not to shift dramatically from the pretrained local minima, provided that the distribution of $\mathcal{D}_\textsc{Finetune}$ does not differ greatly from $\mathcal{D}_\textsc{Train}$. In this experiment, we input a batch of 256 samples from $\mathcal{D}_\textsc{Train}$ to the training loop after every two batches of fine-tuning samples. We choose a learning rate of $5\times 10^{-4}$. Single-label scheme is used in this experiment.

Figure~\ref{fig:resnet_finetune_mix} and Supplementary Information Figure 3 represent the trigger accuracies when fine-tuning with or without training data blending, for ResNet-18 and ViT respectively. For ResNet, it is evident that mixing original training data during fine-tuning improves watermark persistence significantly in the case of noised trigger and unrelated triggers, however, the trend still wiggles dramatically. Data blending seems to be much more beneficial to EWE, as compared with other schemes, regardless of the trigger types. However, in the context of ViT, data blending does not seem to help and in some cases the watermark accuracy drops even faster than the accuracy during normal fine-tuning.

\subsubsection{Retraining in the context of model extraction attack}

\textbf{Model extraction.} Besides fine-tuning attack, we extend our study to model extraction, which is another common attack targeting the confidentiality of ML models. In this attack, the adversary tries to replicate the victim model by training their own model that copies the victim model's behavior on a specific set of outputs. With regard to watermarking, it is very likely that the watermark footprint cannot be retained in an extracted model. Jia et al.~\cite{jiaEntangledWatermarksDefense2021} proposed entangled watermark embedding (EWE) as a watermark scheme resistant to extraction attack by enforcing entanglement between trigger samples and the main task samples. With this technique, an adversary who attempts to remove watermark from extracted model has to sacrifice model performance on the main classification task.

In this experiment, the threat model assumes that the adversary has knowledge about the victim model's architecture as well as access to the data on which the victim model is trained. The extract model is trained on samples from $\mathcal{D}_\textsc{Train}$, where the labels are predicted hard output labels from the victim model. Here, we only experiment with single-label setting since it is easier for models to learn. Supplementary Information Table III shows the watermark accuracy after model extraction. In terms of ResNet, most watermark schemes can retain acceptable accuracy for unrelated triggers. EWE outperforms all other schemes, with high watermark accuracy for noise (65.50\%), unrelated (69.00\%) and FGSM (65.00\%) trigger set. Regarding ViT, we only implemented three schemes Adi, Certified and APP due to the limitations mentioned previously. This time, APP gives a surprisingly high accuracy for unrelated (96.50\%) and FGSM (73.50\%) triggers, while the accuracy for noise triggers is still pretty good at 47\%.

\textbf{Retraining for watermark restoration.} After the attack, we retrain the extracted model on $\mathcal{D}_\textsc{Train}$ with ground-truth labels. Supplementary Information Figure 4 illustrates the watermark performance within 30 epochs of retraining. From the graphs, it is clear that retraining does not improve watermark performance. One hypothesis for this behavior is the extracted model parameters are learned in a different geometry compared to the victim model parameters. Therefore, retraining alone is not able to help parameters move close the local minima of the victim model. In Supplementary Information Figure 5, we visualize the loss landscape in model extraction scenario. It can be seen that the trajectories of retraining do not turn as sharply as in the case of fine-tuning attack shown. This concludes that in the event of extraction attack, retraining does not ensure that model parameters will be guided towards desirable local minima.

\subsection{Key Findings}
Here, we summarize the key observations from our experiments:
\begin{itemize}
    \item \textbf{Watermark persistence during fine-tuning} - the trigger accuracies of Adi~\cite{adiTurningYourWeakness2018}, ROWBACK~\cite{chattopadhyayROWBACKRObustWatermarking2021}, Certified~\cite{bansalCertifiedNeuralNetwork2022}, EWE~\cite{jiaEntangledWatermarksDefense2021} and APP~\cite{ganRobustModelWatermark2023} all decrease during fine-tuning. Our experiments show that no watermark scheme among these is consistently superior to the others in terms of trigger accuracy across various trigger set types, fine-tuning learning rates and labeling schemes.
    \item \textbf{Labeling schemes and trigger types} - our results demonstrate that single-label scheme is more preferable in terms of watermark persistence. A majority of cases show that assigning a fixed label for trigger samples makes it easier for NNs to retain watermark during fine-tuning. One hypothesis for this is that when using multiple labels for trigger set, it makes the NN use more capacity to learn how to distinguish between many ``abnormal'' features. Furthermore, a greater number of classes means fewer trigger samples in each class, which leads to inadequate trigger samples needed for a model to be trained on particular classes. In terms of trigger types, unrelated samples make it easier for models to restore the watermark behavior, while text-overlaid samples cannot be easily restored by solely retraining with $\mathcal{D}_\textsc{Train}$.
    \item \textbf{Retraining with original training data saves watermark after fine-tuning attack} - if model parameters do not drastically shift from their local minima after fine-tuning, there are chances that retraining model with $\mathcal{D}_\textsc{Train}$ brings the watermark back, though the watermark might not be restored completely. In the scope of this paper, we quantify the level of parameters deviation by learning rate. As shown in our experiments, the trigger accuracy can be reinstated to up to $100\%$ depending on various conditions.
    \item \textbf{However, retraining does not help with model extraction} - our empirical results show that retraining is not effective in reinforcing watermark in extracted models. This stems from the fact that the extracted model is learned in a different geometry from the original model parameters.
    \item \textbf{Incorporating original $\mathcal{D}_\textsc{Train}$ in fine-tuning improves watermark erosion in certain cases} - our empirical outcomes show a slight improvement in reducing watermark erosion. For ResNet-18 (Figure~\ref{fig:resnet_finetune_mix}), there is a slight advantage of data blending that is observable with noised and unrelated trigger samples. The benefit becomes very noticeable in the case of EWE. In contrast, as regards ViT (Supplementary Information Figure 3), blending training samples into fine-tuning data does not help alleviate watermark diminishing.
\end{itemize}

\section{Conclusion}\label{section:conclusion}
In this work, we extensively evaluate the persistence of backdoor-based watermark schemes as well as explore a new perspective of watermark restoration. Our empirical study shows that original training data is useful in restoration of watermark footprint which was previously diminished during fine-tuning, provided that appropriate trigger set types are used and the model parameters do not dramatically shift out of their basins of attraction. Besides quantitative evaluation, we visually demonstrate the optimization trend of model parameters via loss landscape geometry. It can be found from our study that by exposing the model to training data after fine-tuning, the learning trajectory interestingly moves back to the local optimum that yields high trigger accuracy, depending on the trigger set type, model type and learning rate during fine-tuning. In our final experiment, we blend the training samples with the fine-tuning data to mitigate the effect of watermark vanishing during fine-tuning. Our experimental results show an improvement over normal fine-tuning in some certain cases. This behavior is notable and serves as a good starting point for more in-depth explorations in future studies.

We are hopeful that this study contributes a new research direction to enhance the robustness and persistence of DNN watermarks. This could facilitate a more data-centric approach when it comes to protect the privacy of machine learning models. We also believe that this work serves as an open problem for future works that dive deep into the theoretical aspect of optimization to enhance the persistence and robustness of neural network watermarks.

\section*{Acknowledgement}
This work is supported by Nanyang Technological University-Desay SV Research Program under Grant 2018-0980.


%



\ifCLASSOPTIONcaptionsoff
  \newpage
\fi



\bibliographystyle{IEEEtran}
\bibliography{bibtex/bib/IEEEabrv, bibtex/bib/references}
\end{document}


%
\title{Persistence of Backdoor-based Watermarks for Neural Networks: A Comprehensive Evaluation\\ \bigskip \LARGE Supplementary Information}
\markboth{IEEE Transactions on Neural Networks and Learning Systems}%
{Ngo \MakeLowercase{\textit{et al.}}: Persistence of Backdoor-based Watermarks for Neural Networks: A Comprehensive Evaluation}
%



\maketitle


\section{Fine-tuning strategy}
\centerline{\begin{minipage}{0.7\linewidth}
\begin{algorithm}[H]
\KwData{watermarked model $f_\theta$, training samples $\mathcal{D}_\textsc{Train}$ with batch size $B_\textsc{t}$, fine-tuning samples $\mathcal{D}_\textsc{Finetune}$ with batch size $B_\textsc{f}$, number of epochs $N$, training samples are mixed after $M$ batches}
\tcp{$\mathcal{D}_\textsc{Train}$, $\mathcal{D}_\textsc{Finetune}$ are shuffled per epoch}
$numBatch=length(\mathcal{D}_\textsc{Finetune})/B_F$\;
\For{$epoch\leftarrow 1$ \KwTo $N$}{
    \For{$i\leftarrow 1$ \KwTo $numBatch$}{
        $X_\textsc{f}, y_\textsc{f}\leftarrow \mathcal{D}_\textsc{Finetune}[i:i+B_\textsc{f}]$\;
        $\textsc{Train}(f_\theta, X_\textsc{f},y_\textsc{f})$\;
        \If{$i \mod M = 0$}{
            $X_\textsc{t}, y_\textsc{t} \leftarrow \mathcal{D}_\textsc{Train}[(i/M):(i/M+B_\textsc{t})]$\;
            $\textsc{Train}(f_\theta, X_\textsc{t},y_\textsc{t})$\;
        }   
    }
}
 \SetKwInOut{Output}{Output}
 \Output{$f_\theta$}
 \caption{Fine-tuning strategy}
 \label{alg:finetuning}
\end{algorithm}
\end{minipage}}

\section{Multi-labeling algorithm}

\centerline{\begin{minipage}{0.7\linewidth}
\begin{algorithm}[H]
\KwData{non-marked pretrained model $\mathcal{M}$, trigger set $\mathcal{D}_{\text{WM}}$}
 \For{$x,y \in \mathcal{D}_{\text{WM}}$}{
    \eIf{$TriggerType$ is FGSM}{
    $x_{adv}, y_{adv} = \textsc{FGSM}(x, \mathcal{M})$\;
    $y \overset{{\scriptscriptstyle \operatorname{R}}}{\leftarrow} \{y_t\in AllClasses \mid y_t \neq y \wedge y_t \neq y_{adv}\}$\;
    }{
    $y\leftarrow (y+1)\mod{NumClasses}$\;
    }
 }
 \caption{Multi-labeling scheme for trigger data}
 \label{alg:multilabel}
\end{algorithm}
\end{minipage}}

\clearpage
\section{Hyper-parameters}
\begin{table}[htbp]
    \centering
\begin{tabular}{llcccccccc}
        \toprule
         \multirow{2}{*}{Stage} & \multirow{2}{*}{Hyperparam} & \multicolumn{5}{c}{ResNet-18} & \multicolumn{3}{c}{ViT-s} \\
         \cmidrule(lr){3-7}\cmidrule(lr){8-10}
         & & Adi & ROWBACK & Certified & EWE & APP & Adi & Certified & APP\\
         \midrule
         \multirow{9}{*}{Pretrain} & LR start & $1\mathrm{e}{-3}$ &$1\mathrm{e}{-3}$ &$1\mathrm{e}{-3}$ & $1\mathrm{e}{-3}$& $1\mathrm{e}{-3}$&$1\mathrm{e}{-3}$ & $1\mathrm{e}{-3}$& $1\mathrm{e}{-3}$\\
          & LR end &$1\mathrm{e}{-5}$ & $1\mathrm{e}{-5}$&$1\mathrm{e}{-5}$ & $1\mathrm{e}{-5}$ &$1\mathrm{e}{-5}$ & $1\mathrm{e}{-5}$&$1\mathrm{e}{-5}$ &$1\mathrm{e}{-5}$ \\
          & Scheduler & cosine & cosine& cosine& cosine& cosine& cosine& cosine & cosine\\
          & Optim & Adam &Adam & Adam& Adam& Adam& Adam&Adam &Adam \\
          & W. Decay & $1\mathrm{e}{-4}$& $1\mathrm{e}{-4}$& $1\mathrm{e}{-4}$&$1\mathrm{e}{-4}$ & $1\mathrm{e}{-4}$& $1\mathrm{e}{-4}$& $1\mathrm{e}{-4}$& $1\mathrm{e}{-4}$\\
          & Epochs & 50&50 &50 & 50& 50&100 &100 & 100\\
\bottomrule
    \end{tabular}
    \caption{Hyper-parameters for each training stage}
    \label{tab:hyperparams}
\end{table}

\clearpage
\section{Model performance after watermark embedding}

    \begin{table}[htbp]
    \centering
    \begin{tabular}{cccccccccc}
        \toprule
        \multirow{6}{*}{Watermark} & \multirow{6}{*}{Trigger} & \multicolumn{8}{c}{Accuracy (\%)}\\
        \cmidrule(lr){3-10}
         & & \multicolumn{4}{c}{Single-label} & \multicolumn{4}{c}{Multi-label}\\
         \cmidrule(lr){3-6}\cmidrule(lr){7-10}
         & & \multicolumn{2}{c}{ResNet} & \multicolumn{2}{c}{ViT} & \multicolumn{2}{c}{ResNet} & \multicolumn{2}{c}{ViT}\\
         \cmidrule(lr){3-4}\cmidrule(lr){5-6}\cmidrule(lr){7-8}\cmidrule(lr){9-10}
        & & Test & WM & Test & WM & Test & WM & Test & WM\\
        \midrule
        \multirow{4}{*}{Adi} & Noise  & 87.49 & 100.00 & 79.48 & 100.00 & 87.85 & 100.00 & 80.19 & 100.00 \\
        {} & Content & 87.80 & 100.00 & 79.50 & 100.00 & 87.65 & 100.00 & 79.32 & 100.00\\
        {} & Unrelated & 87.54 & 100.00 & 78.36 & 100.00 & 87.47 & 100.00 & 78.59 & 100.00\\
        {} & FGSM & 87.45 & 100.00 & 78.78 & 100.00 & 88.06 & 100.00 & 78.87 & 100.00\\
\bottomrule
    \end{tabular}
\caption{Model performance after initial training and watermark embedding}
\label{table:inittrain}
\end{table}

\section{Model extraction accuracy}

\begin{table}[htbp]
    \centering
    \begin{tabularx}{0.6\linewidth}{XXccccc}
        \toprule
        \multirow{2}{*}{Model} & \multirow{2}{*}{Trigger} & \multicolumn{5}{c}{WM accuracy (\%)}\\
        \cmidrule{3-7}
        & & Adi & ROWBACK & Certified & EWE & APP\\
        \midrule
        \multirow{4}{*}{ResNet} & Noise & 0.00 & 0.00 & 0.00 & 65.50 & 0.00\\
        & Content & 1.50 & 6.50 & 4.50 & 8.50 & 3.50\\
        & Unrelated & 18.00 & 42.00 & 30.50 & 69.00 & 40.50\\
        & FGSM & 19.50 & 26.00 & 14.00 & 65.00 & 12.50\\
        \midrule
        \multirow{4}{*}{ViT} & Noise & 4.50 & --- & 30.00 & --- & 47.00 \\
        & Content & 4.00 & --- & 3.00 & --- & 7.50 \\
        & Unrelated & 27.50 & --- & 87.50 & --- & 96.50 \\
        & FGSM & 26.50 & --- & 38.50 & --- & 73.50 \\
        \bottomrule
    \end{tabularx}
\caption{Watermark accuracy after extraction}
\label{table:extract}
\end{table}

\clearpage
\section{Figures}

\begin{figure}[h]
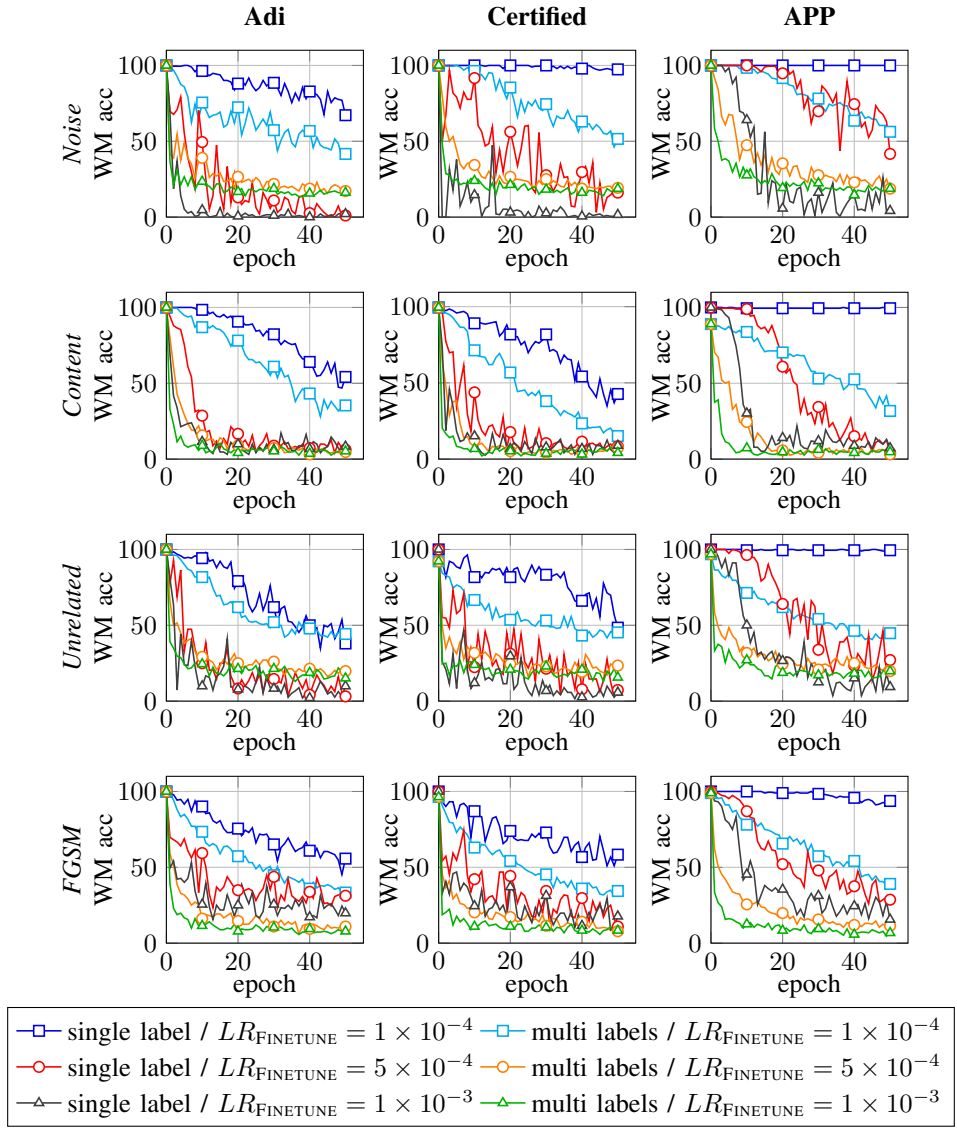

\centering
%
\\
\ref{vit_finetune}
\caption{\textit{Trigger accuracy during fine-tuning of ViT models}}
\label{fig:vit_finetune}
\end{figure}

\begin{figure}[h]
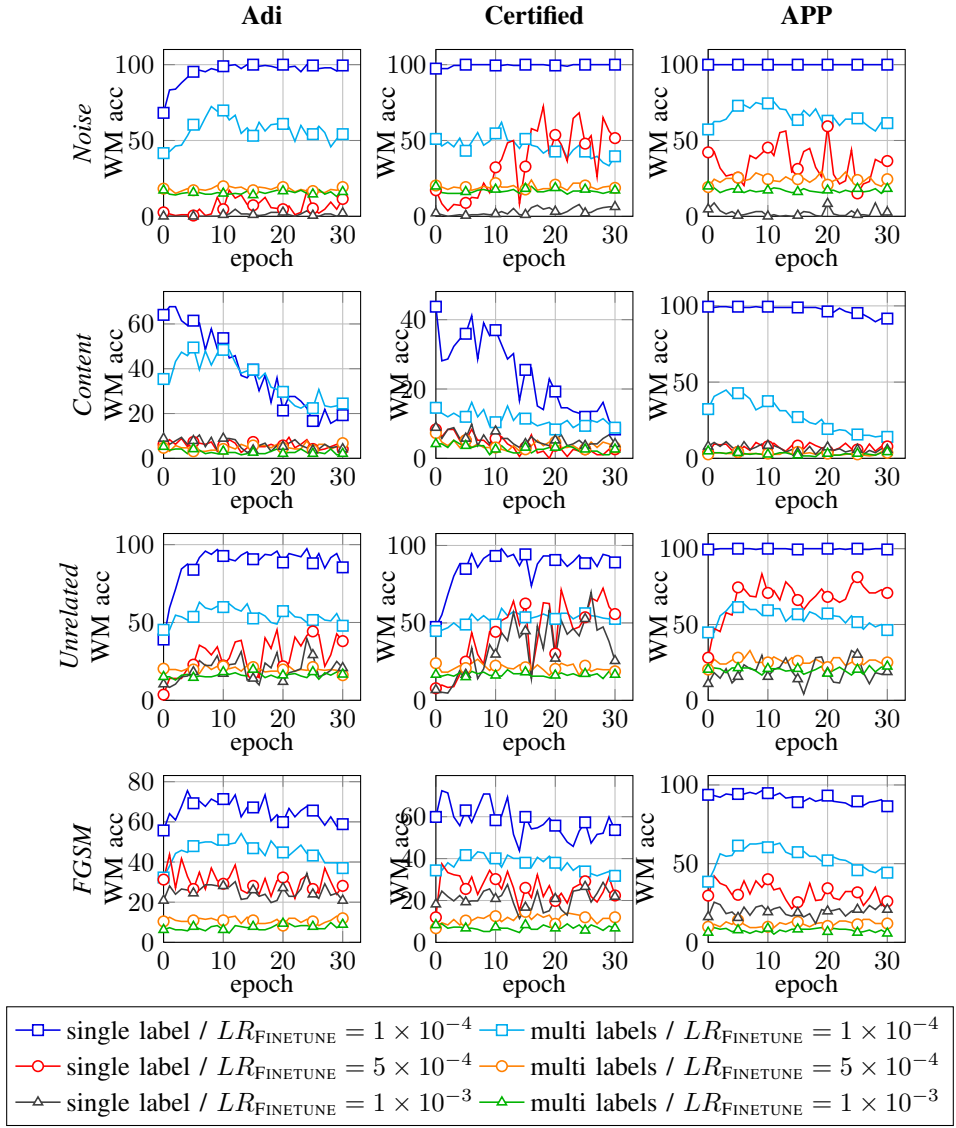

\centering
%
\\
\ref{vit_retrain}
\caption{\textit{Trigger accuracy during retraining of ViT models}}
\label{fig:vit_retrain}
\end{figure}

\begin{figure}[h]
\centering
%
\\
\ref{vit_finetune_mix}
\caption{\textit{Comparison of trigger accuracies between mixing and without mixing of training data $\mathcal{D}_\textsc{Train}$} (ViT)}
\label{fig:vit_finetune_mix}
\end{figure}







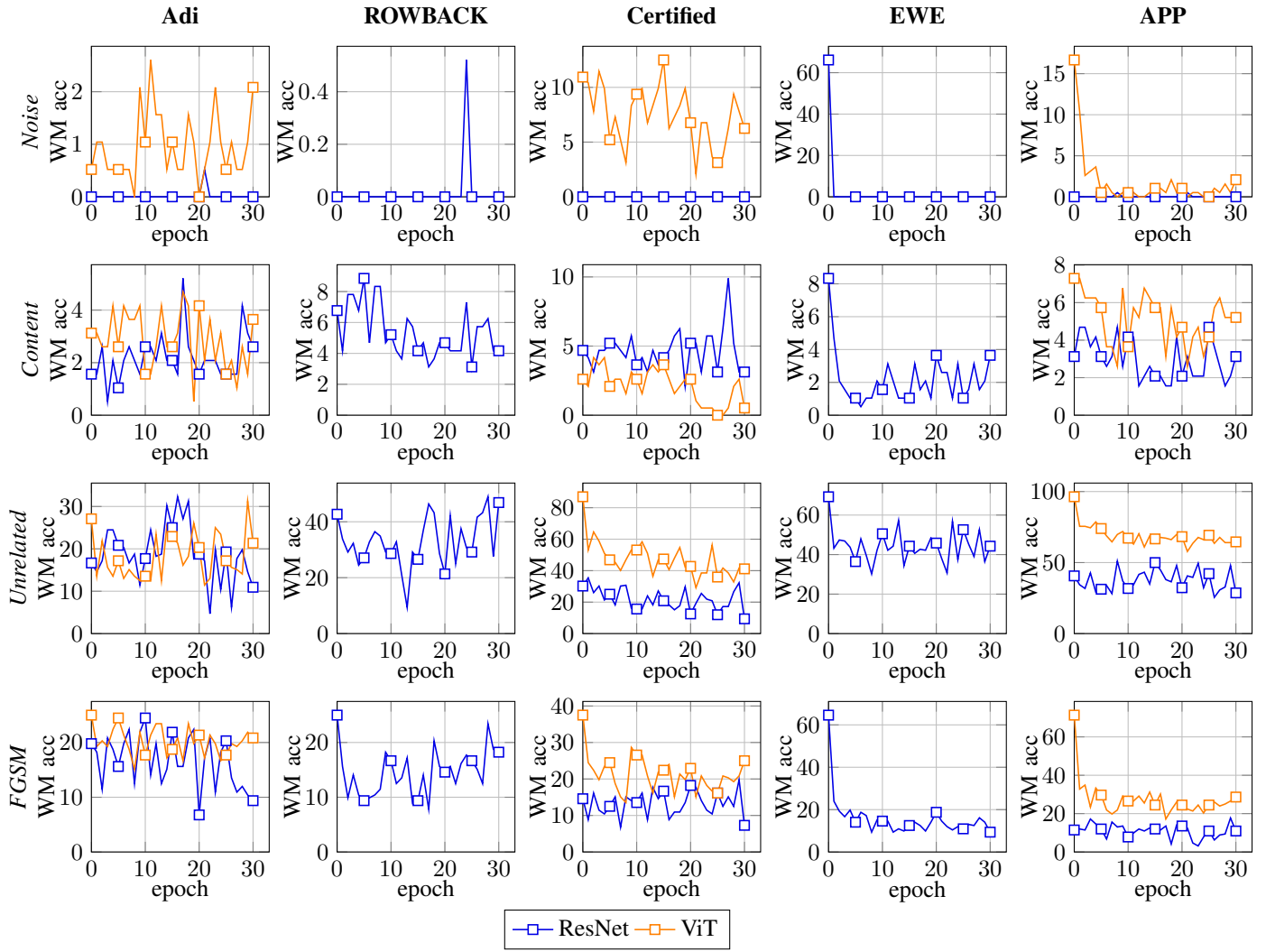
\begin{figure}[h]
\centering
\begin{tikzpicture}
    \begin{groupplot}[group style={group size=5 by 4},height=3.8cm,width=4.2cm]
    \nextgroupplot[
        legend to name=extract_retrain,
        legend entries={ResNet, ViT},
        legend columns=2,
        title={\textbf{Adi}},
        grid=major,
        ylabel style={align=center},
        ylabel shift=-5pt,
        ylabel={\textit{Noise}\\WM acc},
        xlabel shift=-5pt, xlabel={epoch},
        ymin=0,
        xmin=0
    ]
        \addplot[DarkBlue, every mark/.append style={solid, fill=white},mark=square*,mark repeat=5,mark size=2pt, line width=0.6pt] table[x=epoch,y={adi_noise_resnet_cifar10},col sep=comma]{"csv/wandb_extract_retrain_acc.csv"};
        \addplot[orange,every mark/.append style={solid, fill=white},mark=square*,mark repeat=5,mark size=2pt, line width=0.6pt] table[x=epoch,y={adi_noise_vit_cifar10},col sep=comma]{"csv/wandb_extract_retrain_acc.csv"};
    \nextgroupplot[
        title={\textbf{ROWBACK}},
        grid=major,
        ylabel shift=-5pt, ylabel={WM acc},
        xlabel shift=-5pt, xlabel={epoch},
        ymin=0,
        xmin=0
    ]
        \addplot[DarkBlue, every mark/.append style={solid, fill=white},mark=square*,mark repeat=5,mark size=2pt, line width=0.6pt] table[x=epoch,y={rowback_noise_resnet_cifar10},col sep=comma]{"csv/wandb_extract_retrain_acc.csv"};
     \nextgroupplot[
        title={\textbf{Certified}},
        grid=major,
        ymin=0,
        xmin=0,
        ylabel shift=-5pt, ylabel={WM acc},
        xlabel shift=-5pt, xlabel={epoch},
     ]
        \addplot[DarkBlue, every mark/.append style={solid, fill=white},mark=square*,mark repeat=5,mark size=2pt, line width=0.6pt] table[x=epoch,y={certified_noise_resnet_cifar10},col sep=comma]{"csv/wandb_extract_retrain_acc.csv"};
        \addplot[orange,every mark/.append style={solid, fill=white},mark=square*,mark repeat=5,mark size=2pt, line width=0.6pt] table[x=epoch,y={certified_noise_vit_cifar10},col sep=comma]{"csv/wandb_extract_retrain_acc.csv"};
     \nextgroupplot[
        title={\textbf{EWE}},
        grid=major,
        ymin=0,
        xmin=0,
        ylabel shift=-5pt, ylabel={WM acc},
        xlabel shift=-5pt, xlabel={epoch},
     ]
        \addplot[DarkBlue, every mark/.append style={solid, fill=white},mark=square*,mark repeat=5,mark size=2pt, line width=0.6pt] table[x=epoch,y={ewe_noise_resnet_cifar10},col sep=comma]{"csv/wandb_extract_retrain_acc.csv"};
    \nextgroupplot[
        title={\textbf{APP}},
        grid=major,
        ylabel style={align=center},
        ylabel shift=-5pt, ylabel={WM acc},
        xlabel shift=-5pt, xlabel={epoch},
        ymin=0,
        xmin=0
    ]
        \addplot[DarkBlue, every mark/.append style={solid, fill=white},mark=square*,mark repeat=5,mark size=2pt, line width=0.6pt] table[x=epoch,y={app_noise_resnet_cifar10},col sep=comma]{"csv/wandb_extract_retrain_acc.csv"};
        \addplot[orange,every mark/.append style={solid, fill=white},mark=square*,mark repeat=5,mark size=2pt, line width=0.6pt] table[x=epoch,y={app_noise_vit_cifar10},col sep=comma]{"csv/wandb_extract_retrain_acc.csv"};
    \nextgroupplot[
        grid=major,
        ylabel style={align=center},
        ylabel shift=-5pt,
        ylabel={\textit{Content}\\WM acc},
        xlabel shift=-5pt, xlabel={epoch},
        ymin=0,
        xmin=0
    ]
        \addplot[DarkBlue, every mark/.append style={solid, fill=white},mark=square*,mark repeat=5,mark size=2pt, line width=0.6pt] table[x=epoch,y={adi_textoverlay_resnet_cifar10},col sep=comma]{"csv/wandb_extract_retrain_acc.csv"};
        \addplot[orange,every mark/.append style={solid, fill=white},mark=square*,mark repeat=5,mark size=2pt, line width=0.6pt] table[x=epoch,y={adi_textoverlay_vit_cifar10},col sep=comma]{"csv/wandb_extract_retrain_acc.csv"};
     \nextgroupplot[
        grid=major,
        ylabel style={align=center},
        ylabel shift=-5pt,
        ylabel={WM acc},
        xlabel shift=-5pt, xlabel={epoch},
        ymin=0,
        xmin=0
     ]
        \addplot[DarkBlue, every mark/.append style={solid, fill=white},mark=square*,mark repeat=5,mark size=2pt, line width=0.6pt] table[x=epoch,y={rowback_textoverlay_resnet_cifar10},col sep=comma]{"csv/wandb_extract_retrain_acc.csv"};
     \nextgroupplot[
        grid=major,
        ylabel style={align=center},
        ylabel shift=-5pt,
        ylabel={WM acc},
        xlabel shift=-5pt, xlabel={epoch},
        ymin=0,
        xmin=0
     ]
        \addplot[DarkBlue, every mark/.append style={solid, fill=white},mark=square*,mark repeat=5,mark size=2pt, line width=0.6pt] table[x=epoch,y={certified_textoverlay_resnet_cifar10},col sep=comma]{"csv/wandb_extract_retrain_acc.csv"};
        \addplot[orange,every mark/.append style={solid, fill=white},mark=square*,mark repeat=5,mark size=2pt, line width=0.6pt] table[x=epoch,y={certified_textoverlay_vit_cifar10},col sep=comma]{"csv/wandb_extract_retrain_acc.csv"};
        
    \nextgroupplot[
        grid=major,
        ylabel style={align=center},
        ylabel shift=-5pt,
        ylabel={WM acc},
        xlabel shift=-5pt, xlabel={epoch},
        ymin=0,
        xmin=0
    ]
        \addplot[DarkBlue, every mark/.append style={solid, fill=white},mark=square*,mark repeat=5,mark size=2pt, line width=0.6pt] table[x=epoch,y={ewe_textoverlay_resnet_cifar10},col sep=comma]{"csv/wandb_extract_retrain_acc.csv"};
    \nextgroupplot[
        grid=major,
        ylabel style={align=center},
        ylabel shift=-5pt,
        ylabel={WM acc},
        xlabel shift=-5pt, xlabel={epoch},
        ymin=0,
        xmin=0
    ]
        \addplot[DarkBlue, every mark/.append style={solid, fill=white},mark=square*,mark repeat=5,mark size=2pt, line width=0.6pt] table[x=epoch,y={app_textoverlay_resnet_cifar10},col sep=comma]{"csv/wandb_extract_retrain_acc.csv"};
        \addplot[orange,every mark/.append style={solid, fill=white},mark=square*,mark repeat=5,mark size=2pt, line width=0.6pt] table[x=epoch,y={app_textoverlay_vit_cifar10},col sep=comma]{"csv/wandb_extract_retrain_acc.csv"};
     \nextgroupplot[
        grid=major,
        ylabel style={align=center},
        ylabel shift=-5pt,
        ylabel={\textit{Unrelated}\\WM acc},
        xlabel shift=-5pt, xlabel={epoch},
        ymin=0,
        xmin=0
     ]
        \addplot[DarkBlue, every mark/.append style={solid, fill=white},mark=square*,mark repeat=5,mark size=2pt, line width=0.6pt] table[x=epoch,y={adi_unrelated_resnet_cifar10},col sep=comma]{"csv/wandb_extract_retrain_acc.csv"};
        \addplot[orange,every mark/.append style={solid, fill=white},mark=square*,mark repeat=5,mark size=2pt, line width=0.6pt] table[x=epoch,y={adi_unrelated_vit_cifar10},col sep=comma]{"csv/wandb_extract_retrain_acc.csv"};
     \nextgroupplot[
        grid=major,
        ylabel style={align=center},
        ylabel shift=-5pt,
        ylabel={WM acc},
        xlabel shift=-5pt, xlabel={epoch},
        ymin=0,
        xmin=0
     ]
        \addplot[DarkBlue, every mark/.append style={solid, fill=white},mark=square*,mark repeat=5,mark size=2pt, line width=0.6pt] table[x=epoch,y={rowback_unrelated_resnet_cifar10},col sep=comma]{"csv/wandb_extract_retrain_acc.csv"};
    \nextgroupplot[
        grid=major,
        ylabel style={align=center},
        ylabel shift=-5pt,
        ylabel={WM acc},
        xlabel shift=-5pt, xlabel={epoch},
        ymin=0,
        xmin=0
    ]
        \addplot[DarkBlue, every mark/.append style={solid, fill=white},mark=square*,mark repeat=5,mark size=2pt, line width=0.6pt] table[x=epoch,y={certified_unrelated_resnet_cifar10},col sep=comma]{"csv/wandb_extract_retrain_acc.csv"};
        \addplot[orange,every mark/.append style={solid, fill=white},mark=square*,mark repeat=5,mark size=2pt, line width=0.6pt] table[x=epoch,y={certified_unrelated_vit_cifar10},col sep=comma]{"csv/wandb_extract_retrain_acc.csv"};
    \nextgroupplot[
        grid=major,
        ylabel style={align=center},
        ylabel shift=-5pt,
        ylabel={WM acc},
        xlabel shift=-5pt, xlabel={epoch},
        ymin=0,
        xmin=0
    ]
        \addplot[DarkBlue, every mark/.append style={solid, fill=white},mark=square*,mark repeat=5,mark size=2pt, line width=0.6pt] table[x=epoch,y={ewe_unrelated_resnet_cifar10},col sep=comma]{"csv/wandb_extract_retrain_acc.csv"};
     \nextgroupplot[
        grid=major,
        ylabel style={align=center},
        ylabel shift=-10pt,
        ylabel={WM acc},
        xlabel shift=-5pt, xlabel={epoch},
        ymin=0,
        xmin=0
     ]
        \addplot[DarkBlue, every mark/.append style={solid, fill=white},mark=square*,mark repeat=5,mark size=2pt, line width=0.6pt] table[x=epoch,y={app_unrelated_resnet_cifar10},col sep=comma]{"csv/wandb_extract_retrain_acc.csv"};
        \addplot[orange,every mark/.append style={solid, fill=white},mark=square*,mark repeat=5,mark size=2pt, line width=0.6pt] table[x=epoch,y={app_unrelated_vit_cifar10},col sep=comma]{"csv/wandb_extract_retrain_acc.csv"};
     \nextgroupplot[
        grid=major,
        ylabel style={align=center},
        ylabel shift=-5pt,
        ylabel={\textit{FGSM}\\WM acc},
        xlabel shift=-5pt, xlabel={epoch},
        ymin=0,
        xmin=0
     ]
        \addplot[DarkBlue, every mark/.append style={solid, fill=white},mark=square*,mark repeat=5,mark size=2pt, line width=0.6pt] table[x=epoch,y={adi_adv_resnet_cifar10},col sep=comma]{"csv/wandb_extract_retrain_acc.csv"};
        \addplot[orange,every mark/.append style={solid, fill=white},mark=square*,mark repeat=5,mark size=2pt, line width=0.6pt] table[x=epoch,y={adi_adv_vit_cifar10},col sep=comma]{"csv/wandb_extract_retrain_acc.csv"};
    \nextgroupplot[
        grid=major,
        ylabel style={align=center},
        ylabel shift=-5pt,
        ylabel={WM acc},
        xlabel shift=-5pt, xlabel={epoch},
        ymin=0,
        xmin=0
    ]
        \addplot[DarkBlue, every mark/.append style={solid, fill=white},mark=square*,mark repeat=5,mark size=2pt, line width=0.6pt] table[x=epoch,y={rowback_adv_resnet_cifar10},col sep=comma]{"csv/wandb_extract_retrain_acc.csv"};
    \nextgroupplot[
        grid=major,
        ylabel style={align=center},
        ylabel shift=-5pt,
        ylabel={WM acc},
        xlabel shift=-5pt, xlabel={epoch},
        ymin=0,
        xmin=0
    ]
        \addplot[DarkBlue, every mark/.append style={solid, fill=white},mark=square*,mark repeat=5,mark size=2pt, line width=0.6pt] table[x=epoch,y={certified_adv_resnet_cifar10},col sep=comma]{"csv/wandb_extract_retrain_acc.csv"};
        \addplot[orange,every mark/.append style={solid, fill=white},mark=square*,mark repeat=5,mark size=2pt, line width=0.6pt] table[x=epoch,y={certified_adv_vit_cifar10},col sep=comma]{"csv/wandb_extract_retrain_acc.csv"};
     \nextgroupplot[
        grid=major,
        ylabel style={align=center},
        ylabel shift=-5pt,
        ylabel={WM acc},
        xlabel shift=-5pt, xlabel={epoch},
        ymin=0,
        xmin=0
     ]
        \addplot[DarkBlue, every mark/.append style={solid, fill=white},mark=square*,mark repeat=5,mark size=2pt, line width=0.6pt] table[x=epoch,y={ewe_adv_resnet_cifar10},col sep=comma]{"csv/wandb_extract_retrain_acc.csv"};
     \nextgroupplot[
        grid=major,
        ylabel style={align=center},
        ylabel shift=-5pt,
        ylabel={WM acc},
        xlabel shift=-5pt, xlabel={epoch},
        ymin=0,
        xmin=0
     ]
        \addplot[DarkBlue, every mark/.append style={solid, fill=white},mark=square*,mark repeat=5,mark size=2pt, line width=0.6pt] table[x=epoch,y={app_adv_resnet_cifar10},col sep=comma]{"csv/wandb_extract_retrain_acc.csv"};
        \addplot[orange,every mark/.append style={solid, fill=white},mark=square*,mark repeat=5,mark size=2pt, line width=0.6pt] table[x=epoch,y={app_adv_vit_cifar10},col sep=comma]{"csv/wandb_extract_retrain_acc.csv"};
    \end{groupplot}
\end{tikzpicture}\\
\ref{extract_retrain}
\caption{\textit{Trigger accuracy during retraining of extracted models}}
\label{fig:extract_retrain}
\end{figure}

\begin{figure}[h]
    \centering
    \begin{tabular}{@{}m{0.4cm}
    @{}m{\dimexpr0.2\textwidth-0.2cm\relax}
    @{}m{\dimexpr0.2\textwidth-0.2cm\relax}
    @{}m{\dimexpr0.2\textwidth-0.2cm\relax}
    @{}m{\dimexpr0.2\textwidth-0.2cm\relax}
    @{}m{\dimexpr0.2\textwidth-0.2cm\relax}
    @{}}
    \raisebox{-\baselineskip}{\rotatebox{90}{\textit{Noise}}}
    & \begin{subfigure}[b]{\linewidth}
        \centering
        \caption*{\textbf{Adi}}
        \includegraphics[width=\linewidth]{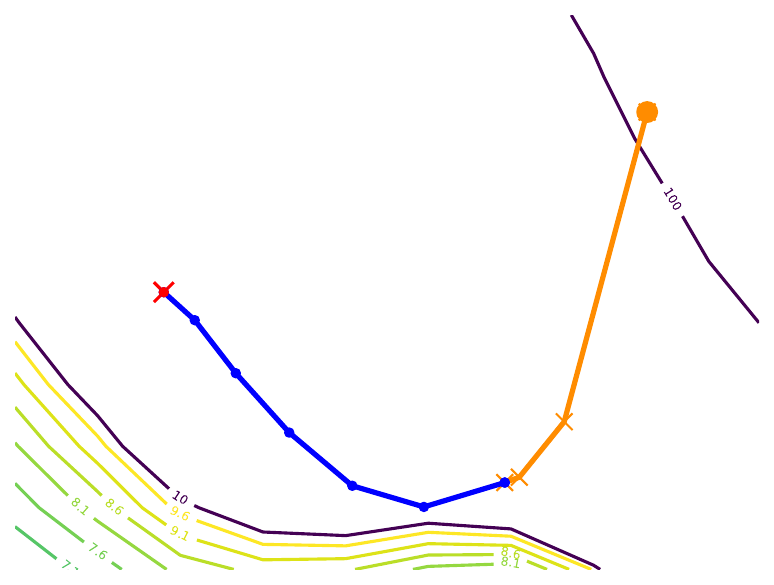} 
    \end{subfigure}  
    & \begin{subfigure}[b]{\linewidth}
        \centering
        \caption*{\textbf{ROWBACK}}
        \includegraphics[width=\linewidth]{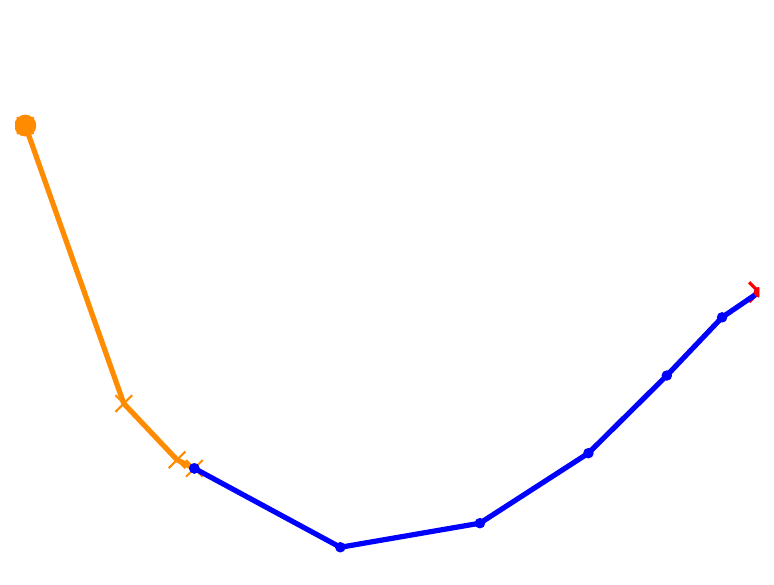} 
    \end{subfigure}
    & \begin{subfigure}[b]{\linewidth}
        \centering
        \caption*{\textbf{Certified}}
        \includegraphics[width=\linewidth]{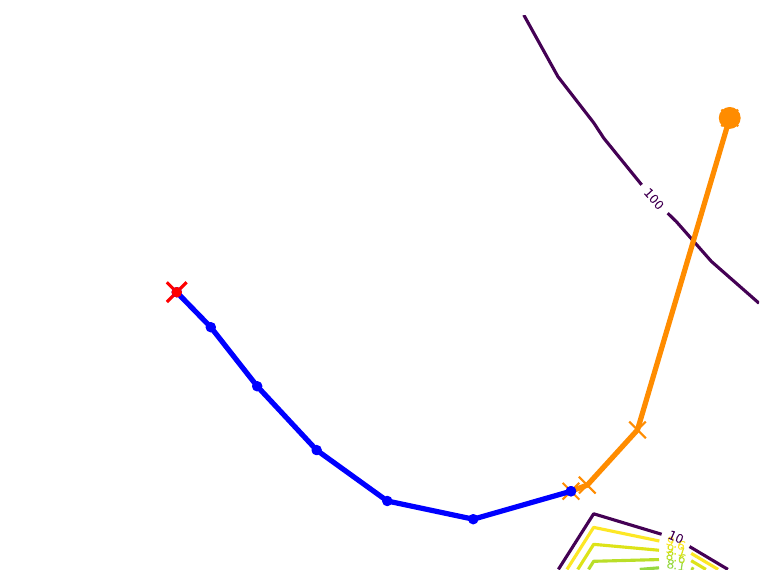} 
        \end{subfigure}  
    & \begin{subfigure}[b]{\linewidth}
        \centering
        \caption*{\textbf{EWE}}
        \includegraphics[width=\linewidth]{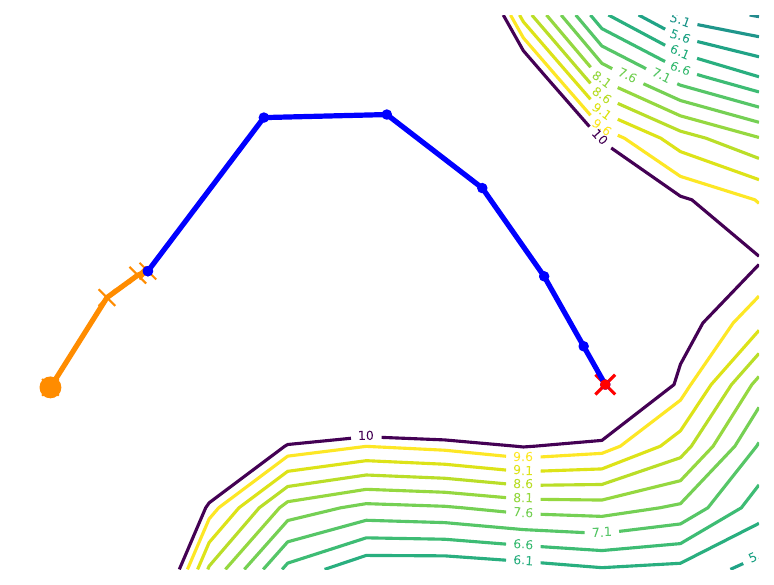} 
    \end{subfigure}
    & \begin{subfigure}[b]{\linewidth}
        \centering
        \caption*{\textbf{APP}}
        \includegraphics[width=\linewidth]{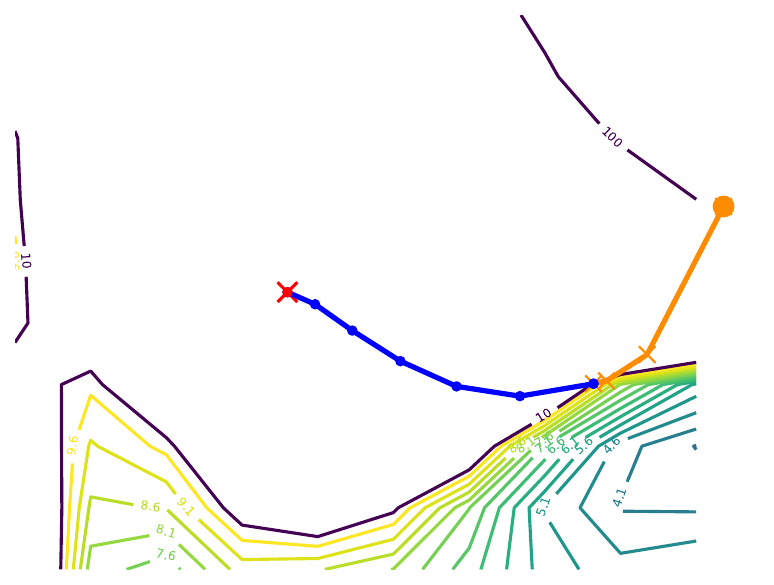} 
    \end{subfigure} \\

    \raisebox{-\baselineskip}{\rotatebox{90}{\textit{Content}}}
    & \begin{subfigure}[b]{\linewidth}
        \centering
        \includegraphics[width=\linewidth]{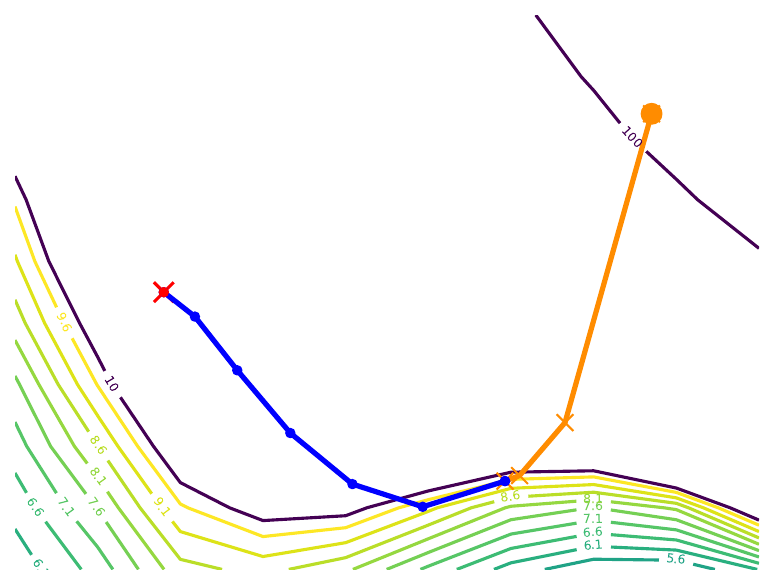} 
    \end{subfigure}  
    & \begin{subfigure}[b]{\linewidth}
        \centering
        \includegraphics[width=\linewidth]{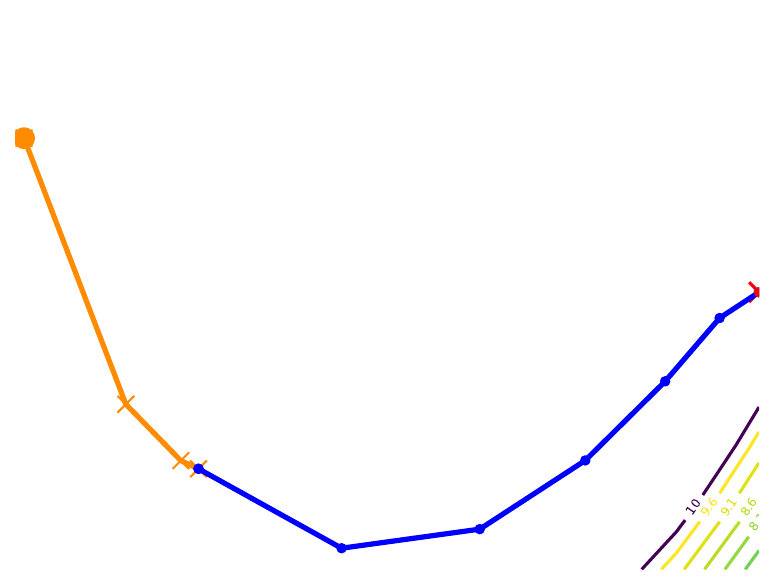} 
    \end{subfigure}
    & \begin{subfigure}[b]{\linewidth}
        \centering
        \includegraphics[width=\linewidth]{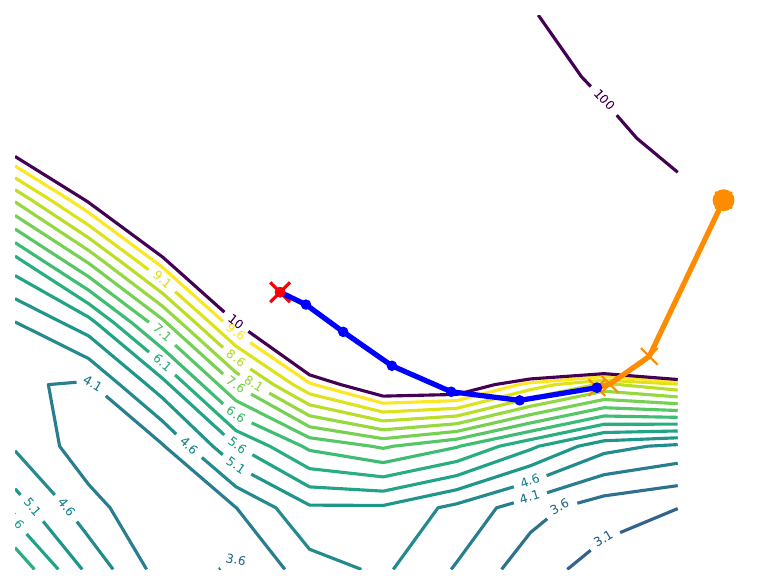} 
        \end{subfigure}  
    & \begin{subfigure}[b]{\linewidth}
        \centering
        \includegraphics[width=\linewidth]{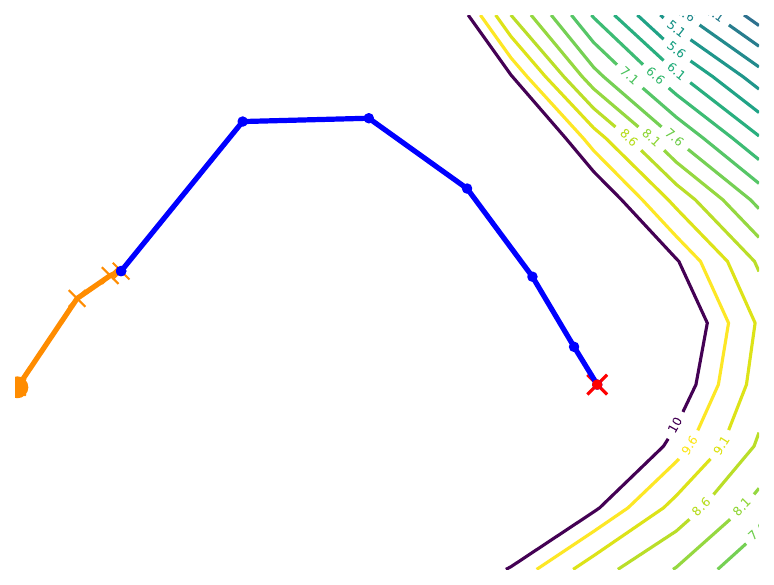} 
    \end{subfigure}
    & \begin{subfigure}[b]{\linewidth}
        \centering
        \includegraphics[width=\linewidth]{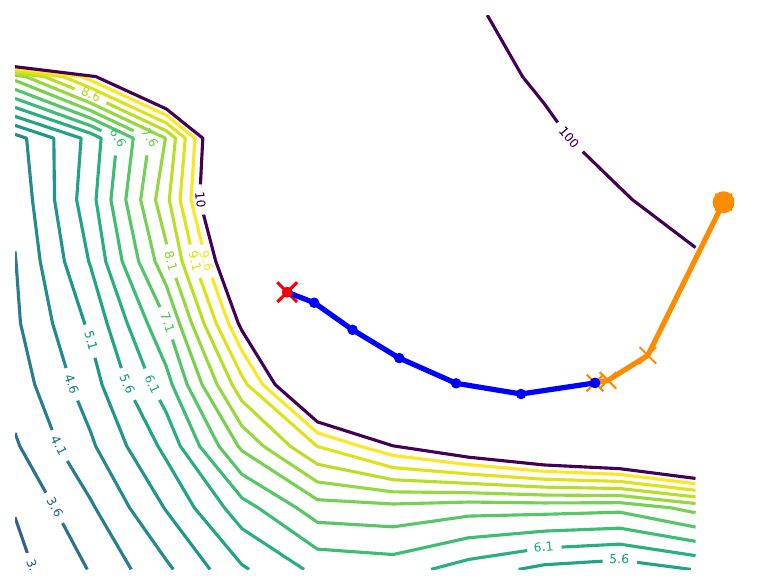} 
    \end{subfigure} \\

    \raisebox{-\baselineskip}{\rotatebox{90}{\textit{Unrelated}}}
    & \begin{subfigure}[b]{\linewidth}
        \centering
        \includegraphics[width=\linewidth]{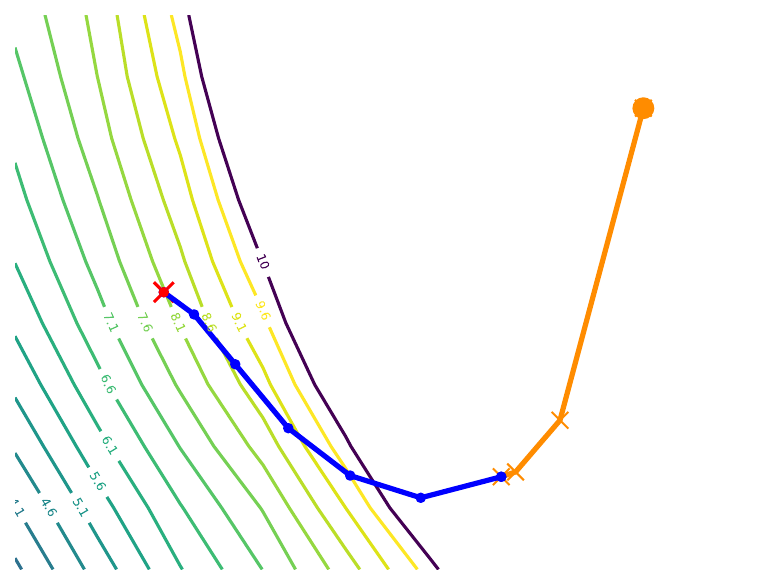} 
    \end{subfigure}  
    & \begin{subfigure}[b]{\linewidth}
        \centering
        \includegraphics[width=\linewidth]{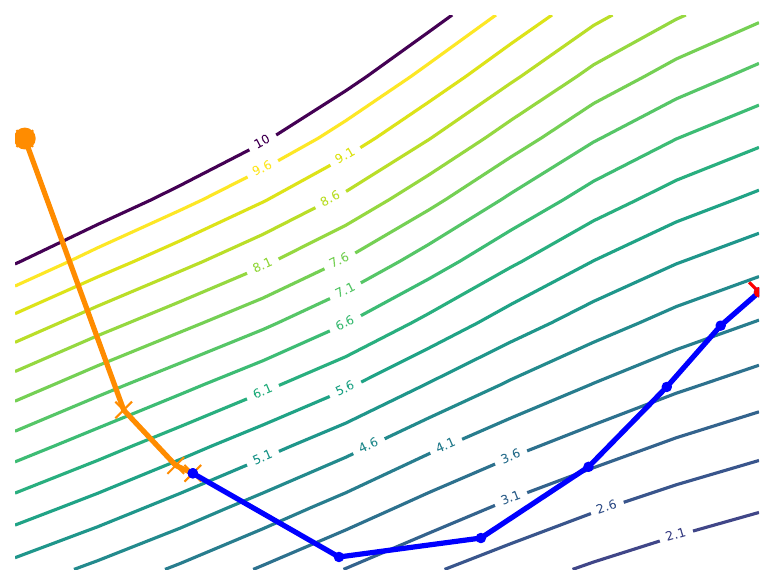} 
    \end{subfigure}
    & \begin{subfigure}[b]{\linewidth}
        \centering
        \includegraphics[width=\linewidth]{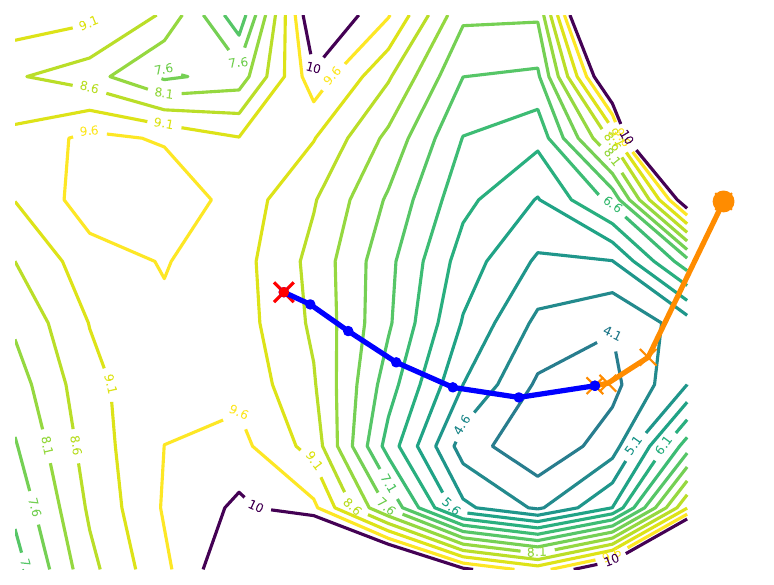} 
        \end{subfigure}  
    & \begin{subfigure}[b]{\linewidth}
        \centering
        \includegraphics[width=\linewidth]{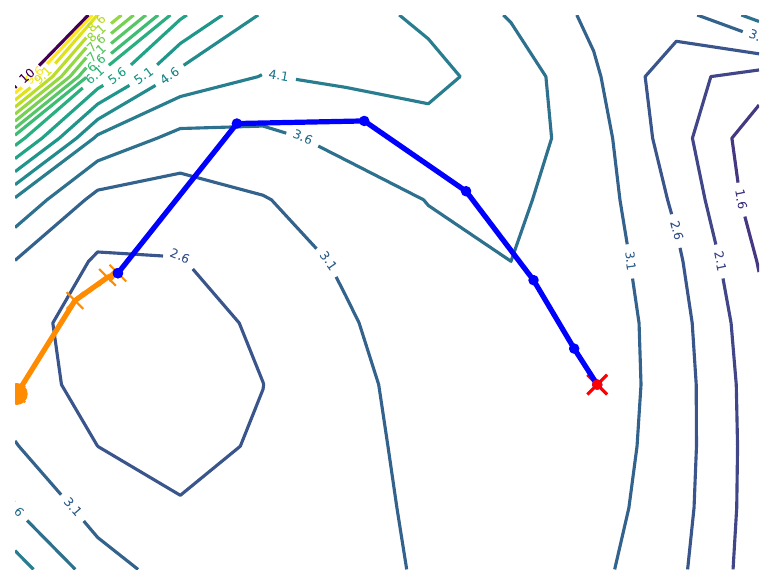} 
    \end{subfigure}
    & \begin{subfigure}[b]{\linewidth}
        \centering
        \includegraphics[width=\linewidth]{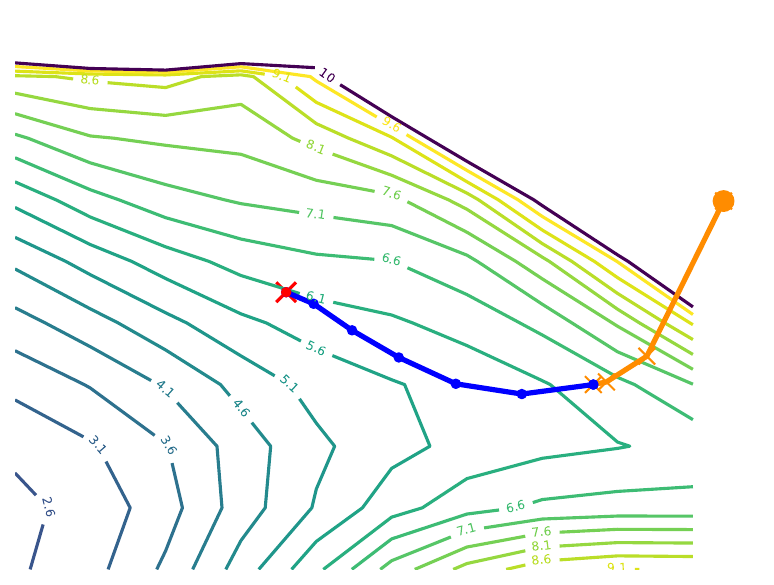} 
    \end{subfigure} \\

    \raisebox{-\baselineskip}{\rotatebox{90}{\textit{FGSM}}}
    & \begin{subfigure}[b]{\linewidth}
        \centering
        \includegraphics[width=\linewidth]{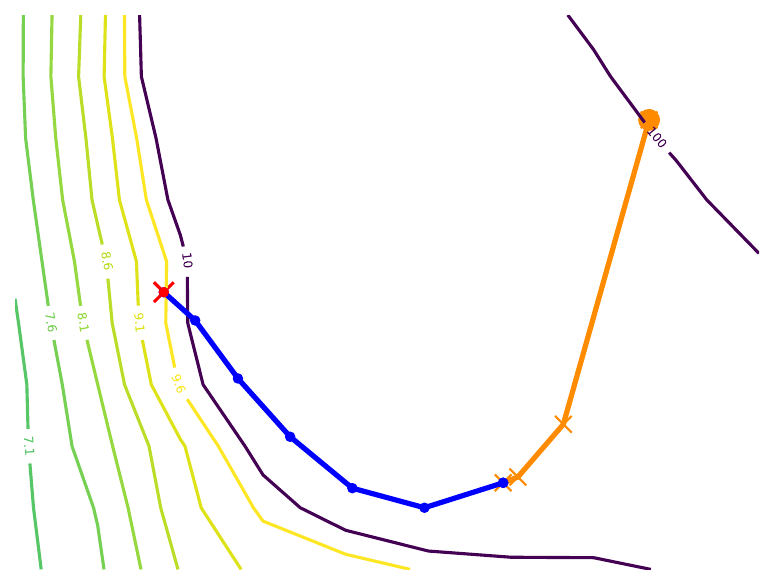} 
    \end{subfigure}  
    & \begin{subfigure}[b]{\linewidth}
        \centering
        \includegraphics[width=\linewidth]{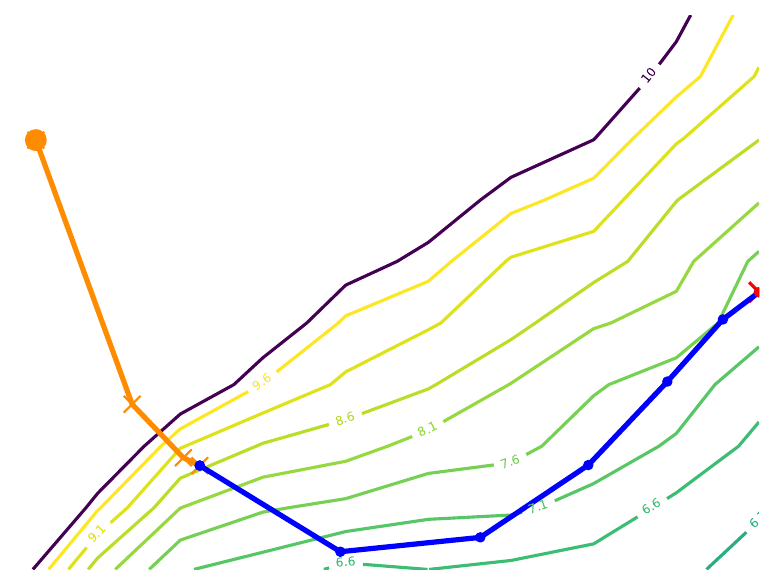} 
    \end{subfigure}
    & \begin{subfigure}[b]{\linewidth}
        \centering
        \includegraphics[width=\linewidth]{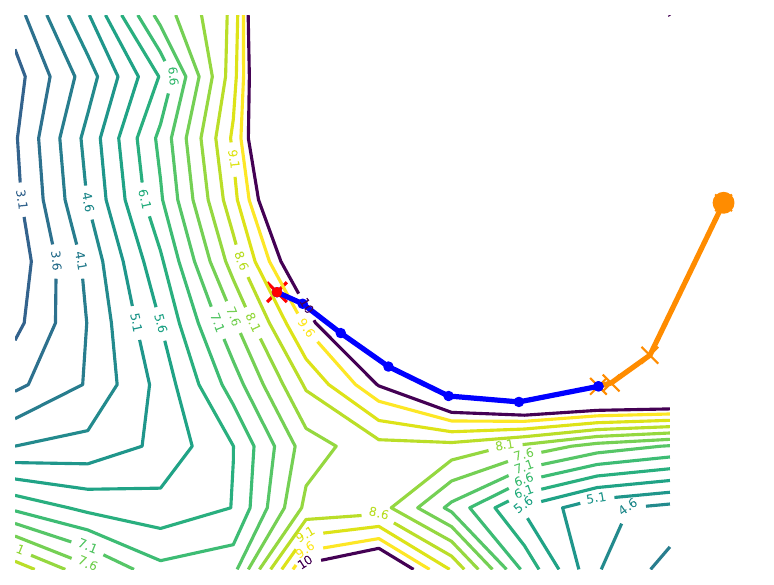} 
        \end{subfigure}  
    & \begin{subfigure}[b]{\linewidth}
        \centering
        \includegraphics[width=\linewidth]{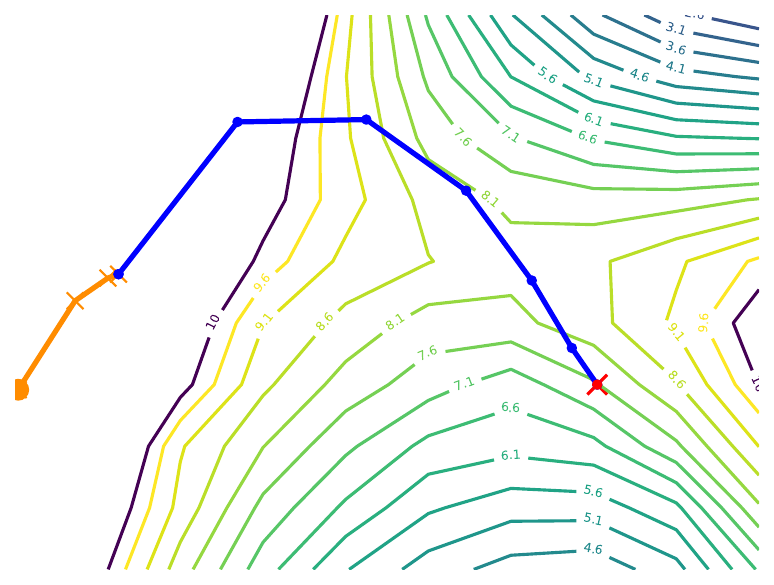} 
    \end{subfigure}
    & \begin{subfigure}[b]{\linewidth}
        \centering
        \includegraphics[width=\linewidth]{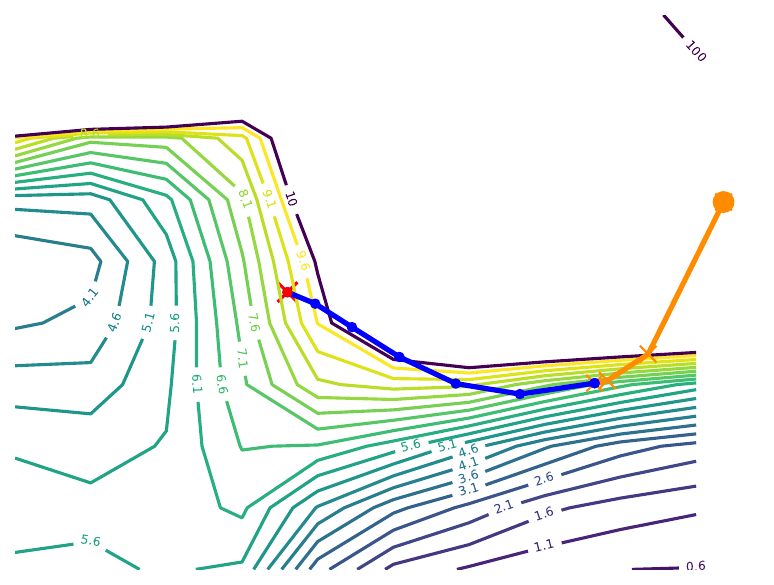} 
    \end{subfigure} \\
  
    \end{tabular}
    \caption{\textit{Loss landscape visualization for model extraction (ResNet)} - The contours illustrate trigger loss, \textcolor{orange}{\textit{orange}} lines depict a few last epochs of extraction phase while \textcolor{blue}{\textit{blue}} lines represent retraining. It can be seen that the trajectories during retraining do not turn as sharply as in fine-tuning attack.}
    \label{fig:contours_extract}
\end{figure}

\ifCLASSOPTIONcaptionsoff
  \newpage
\fi